\documentclass[sigconf]{acmart}
\AtBeginDocument{%
  }

\copyrightyear{2025}
\acmYear{2025}
\setcopyright{acmlicensed}
\acmConference[MM '25]{Proceedings of the 33rd ACM International Conference on Multimedia}{October 27--31, 2025}{Dublin, Ireland}
\acmBooktitle{Proceedings of the 33rd ACM International Conference on Multimedia (MM '25), October 27--31, 2025, Dublin, Ireland}
\acmDOI{10.1145/3746027.3755764}
\acmISBN{979-8-4007-2035-2/2025/10}

\usepackage{multirow} 
\usepackage{tabularray}
\usepackage{bm}
\usepackage{enumitem}
\usepackage{algorithm} 
\usepackage{algpseudocode} 

\begin{document}

\title{Enhancing Multimodal In-Context Learning for Image Classification through Coreset Optimization}

\author{Huiyi Chen}
\orcid{0009-0002-9263-7190}
\affiliation{%
  \institution{Southeast University}
  \department{School of Computer Science and Engineering}
  \city{Nanjing}
  \country{China}}
\email{huiyichen@seu.edu.cn}

\author{Jiawei Peng}
\orcid{0009-0005-1751-239X}
\affiliation{%
  \institution{Southeast University}
  \department{School of Computer Science and Engineering}
  \city{Nanjing}
  \country{China}
}
\email{pengjiawei@seu.edu.cn}

\author{Kaihua Tang}
\orcid{0000-0002-1008-7053}
\affiliation{%
 \institution{Huawei Singapore Research Center}
 \city{Singapore}
 \country{Singapore}}
 \email{tkhchipaomian@gmail.com}

\author{Xin Geng}
\orcid{0000-0001-7729-0622}
\affiliation{%
  \institution{Southeast University}
  \department{School of Computer Science and Engineering}
  \city{Nanjing}
  \country{China}}
  \email{xgeng@seu.edu.cn}

\author{Xu Yang}
\orcid{0000-0002-8276-2679}
\authornote{Corresponding author.}
\affiliation{%
  \institution{Southeast University}
  \department{School of Computer Science and Engineering}
  \city{Nanjing}
  \country{China}}
  \email{xuyang_palm@seu.edu.cn}

\renewcommand{\shortauthors}{Huiyi Chen, Jiawei Peng, Kaihua Tang, Xin Geng, and Xu Yang}

\begin{abstract}
In-context learning (ICL) enables Large Vision-Language Models (LVLMs) to adapt to new tasks without parameter updates, using a few demonstrations from a large support set. However, selecting informative demonstrations leads to high computational and memory costs. While some methods explore selecting a small and representative coreset in the text classification, evaluating all support set samples remains costly, and discarded samples lead to unnecessary information loss. These methods may also be less effective for image classification due to differences in feature spaces.
Given these limitations, we propose Key-based Coreset Optimization (KeCO), a novel framework that leverages untapped data to construct a compact and informative coreset. 
We introduce visual features as keys within the coreset, which serve as the anchor for identifying samples to be updated through different selection strategies. By leveraging untapped samples from the support set, we update the keys of selected coreset samples, enabling the randomly initialized coreset to evolve into a more informative coreset under low computational cost.
Through extensive experiments on coarse-grained and fine-grained image classification benchmarks, we demonstrate that KeCO effectively enhances ICL performance for image classification task, achieving an average improvement of more than 20\%. Notably, we evaluate KeCO under a simulated online scenario, and the strong performance in this scenario highlights the practical value of our framework for resource-constrained real-world scenarios. Our code is provided in: \url{https://github.com/chenyil6/KeCO_Coreset_Optimization}. 
\end{abstract}


\begin{teaserfigure}
  \centering
  \includegraphics[width=0.95\textwidth]{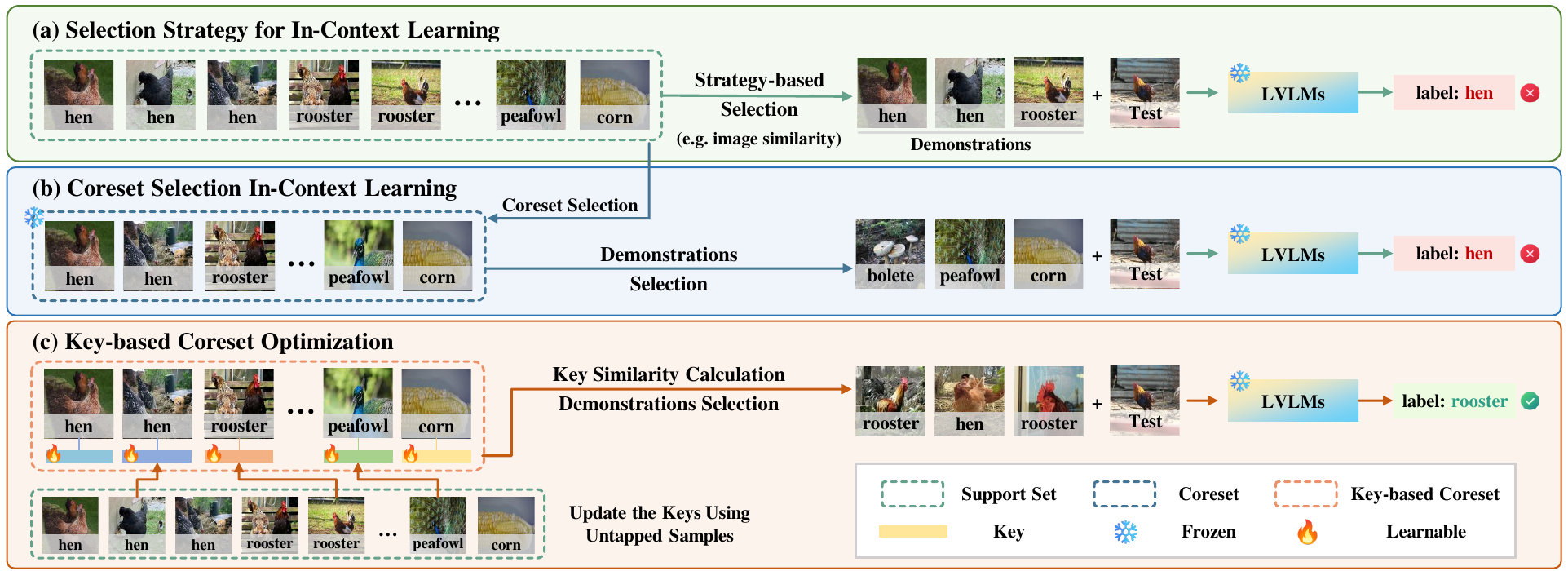}
  \caption{(a) Strategy-Based Selection In-Context Learning: requires storing the full support set and computing similarities between each test input and all support samples before inference. (b) Coreset Selection In-Context Learning: uses a coreset selection strategy to explore an informative subset from the entire support set, reducing the cost of selecting demonstrations. (c) Key-based Coreset Optimization: the untapped samples in the support set are used to update the coreset. Specifically, this involves updating the key of each sample, where the key refers to the visual feature of the sample.}
  \label{fig:overview}
\end{teaserfigure}


\maketitle

\section{Introduction}
In-context learning (ICL) has emerged as a powerful paradigm~\cite{brown2020language,dong2022survey} that enables Large Language Models (LLMs) to solve novel tasks by conditioning on a few in-context examples (ICEs), without additional training. This paradigm is particularly appealing in real-world scenarios where fine-tuning is costly or infeasible. Recently, ICL has been extended to the multi-modal setting, where Large Vision-Language Models (LVLMs) leverage a few image-text pairs as ICEs to perform downstream tasks~\cite{alayrac2022flamingo,li2023mimic,tsimpoukelli2021multimodal,laurenccon2023obelics}. 

Prior studies have shown that ICL performance is highly sensitive to the selection of ICEs~\cite{chang2022data,baldassini2024makes}. Compared to conventional random selection, recent works adopt more effective strategies, such as similarity-based retrieval, to select informative ICEs~\cite{min2022rethinking,zhang2023makes,pan2023context,li2024configure}. However, as illustrated in Figure~\ref{fig:overview} (a), these methods require storing and traversing the entire support set, resulting in significant computational and memory overhead. To mitigate this issue, recent studies in Natural Language Processing (NLP) have explored \textit{coreset} selection techniques, which aim to identify a small yet representative subset of the support set~\cite{sener2017active,li2023finding}. For example, ~\cite{li2023finding} use language model feedback to identify samples whose influence on model predictions surpasses that of other samples in the support set. As shown in Figure~\ref{fig:overview} (b), this approach substantially reduces retrieval costs while maintaining comparable ICL performance.

Despite these advances, coreset-based strategies still face several limitations in ICL. First, identifying a coreset typically involves complex selection mechanisms and time-consuming evaluations of all support set samples. Second, applying sample-level coreset selection to image classification is inherently challenging. Unlike textual data, which is often composed of compact discrete tokens, visual inputs are inherently richer and more diverse~\cite{he2020momentum}. For example, images of a `hen' may vary significantly in pose, background, and other category-irrelevant factors, making it harder to adequately represent the underlying feature space within a fixed small subset. Third, fixed coresets inevitably discard a large portion of the support set. Although these samples are evaluated during selection, the information they contain remains unused in downstream inference, leading to unnecessary information loss.

Given these limitations, we propose a novel coreset construction framework for image classification tasks, termed \textbf{Ke}y-based \textbf{C}oreset \textbf{O}ptimization (KeCO). Instead of relying solely on sample selection, KeCO leverages untapped data to update coreset representations at the feature level, yielding a compact yet informative coreset for effective ICL with LVLMs.
Specifically, KeCO consists of three steps: (1) we begin by randomly sampling an initial coreset from the support set. Although simple, this initialization proves effective when followed by our feature-level refinement procedure. We extract visual features from coreset samples as \textit{keys}, which serve as anchors for subsequent selection and update processes. 
(2) The remaining untapped samples are then used to update the coreset. For each incoming query sample from the untapped set, we retrieve a target sample from the coreset to be updated. We investigate various selection strategies (e.g., diversity-based selection) at this step to examine their impact on coreset optimization.
(3) Once a target sample is selected, its key is updated by incorporating information from the query sample via linear interpolation.

After iteratively running steps (2) and (3) to refine the coreset with untapped samples, we obtain a compact coreset that significantly enhances ICL performance. To further demonstrate its practicality, we extend KeCO to a \textbf{simulated online scenario}, where samples are no longer available all at once but instead arrive in a streaming fashion. KeCO naturally adapts to this setting with minimal adjustments, enabling LVLMs to continuously benefit from incoming samples under constrained memory resources.

We evaluated KeCO on three datasets spanning both coarse-grained and fine-grained image classification datasets. Compared to the fixed coreset, KeCO improves the ICL performance by 20\% for OpenFlamingo-3B (OF-3B)~\cite{awadalla2023openflamingo} and 10\% for IDEFICS-8B (IDE-8B)~\cite{laurenccon2023obelics}.
Remarkably, KeCO even outperforms the ICL using the \textbf{full support set}, which is five times larger than the coreset, by approximately 10\% for OF-3B and 4\% for IDE-8B. We further demonstrate that target sample selection plays a crucial role in our framework, with diversity-based selection yielding the best results. In addition, KeCO achieves larger gains on fine-grained datasets, highlighting its effectiveness in capturing subtle visual distinctions. In the simulated online scenario, KeCO maintains strong and stable performance. For instance, it enables Qwen2-VL~\cite{wang2024qwen2} to exceed the baseline by approximately 5\% on CUB-200. Our main contributions are summarized as follows:

\begin{itemize}[leftmargin=*]
    \item We propose a novel coreset optimization framework, \textbf{KeCO}, which constructs a compact and effective coreset from a large support set. To the best of our knowledge, this is the first attempt to introduce coreset optimization into the ICL paradigm.
    
    \item KeCO leverages untapped data to aggregate category-relevent information into the coreset via feature-level updates. Notably, KeCO achieves strong performance in a simulated online scenario, demonstrating its practical applicability.
    
    \item Extensive experiments show that KeCO significantly boosts ICL performance for image classification task. We further identify that diversity-based selection outperforms other target sample selection strategies, shedding light on better update mechanisms.
\end{itemize}

\section{Related Work}
\subsection{Large Vision-Language Model}
Large Vision-Language Models (LVLMs)~\cite{liang2024survey} are generally built on a vision encoder, a pre-trained Large Language Model (LLM), and an alignment module between the vision encoder and the LLM. The level of performance of these models has started to approach those of LLMs, especially after multimodal instruction tuning~\cite{liu2023visual,huang2023visual,you2022end,Das_2024_ECCV,Hu_2024_CVPR,you2021mrd,you2021self,peng2025lmm}. Although there are numerous LVLMs available, it is important to note that not all of these models support in-context learning (ICL). For example, mPLUG-Owl~\cite{ye2023mplug}, BLIP-2~\cite{li2023blip} and MiniGPT-4~\cite{zhu2023minigpt} lack the capabilities for ICL because they have not undergone dedicated few-shot pre-training and cannot handle the input distribution associated with ICL. In contrast, models like Flamingo~\cite{alayrac2022flamingo}  and IDEFICS~\cite{laurenccon2023obelics} are specifically designed to support this task. In this work, we mainly focus on two open-source models with ICL capabilities (OpenFlamingo~\cite{awadalla2023openflamingo} and IDEFICS~\cite{laurenccon2023obelics}) for our main experiments. Additional experiments are also performed on a high-performance commercial model, Qwen2-VL~\cite{wang2024qwen2}.

\subsection{Multimodal In-context Learning}
ICL enables LLMs to tackle novel tasks by leveraging a few demonstrations from a support set~\cite{brown2020language, dong2022survey}. Building on this success, ICL has been extended to LVLMs~\cite{tsimpoukelli2021multimodal,alayrac2022flamingo,li2023ottermultimodalmodelincontext,laurenccon2024obelics,awadalla2023openflamingo,li2023otterhd,feng2025redefining}.
To further enhance ICL in LVLMs, several works have focused on improving the construction strategies of in-context sequences, such as similarity-based selection~\cite{liu2022makes, pan2023context, li2023unified, min2022rethinking, yang2023exploring} and diversity-based selection~\cite{levy2023diverse, li2024configure}.
Despite its effectiveness, it faces challenges in resource-limited scenarios, requiring large storage and significant computational resources for computation~\cite{yang2023lever,Jiang_2025_CVPR}. Therefore, in the NLP domain, some works focus on selecting samples from an unlabeled pool for annotation to reduce annotation costs~\cite{qian2024sub,mavromatis2023examples}, while others attempt to explore a representative coreset for ICL~\cite{sener2017active,li2023finding}.
However, identifying such a coreset requires complex computation strategies and
time-consuming evaluations of all samples in the support set. 
Additionally, a small coreset may not adequately represent the diverse semantics of the dataset in vision domain, and discarded data from the support set remains untapped. To overcome these limitations, we randomly sample a coreset and then integrate feature-level information from the untapped data to enhance the information contained within this coreset.

\begin{figure*}
  \centering
  \includegraphics[width=0.96\textwidth]{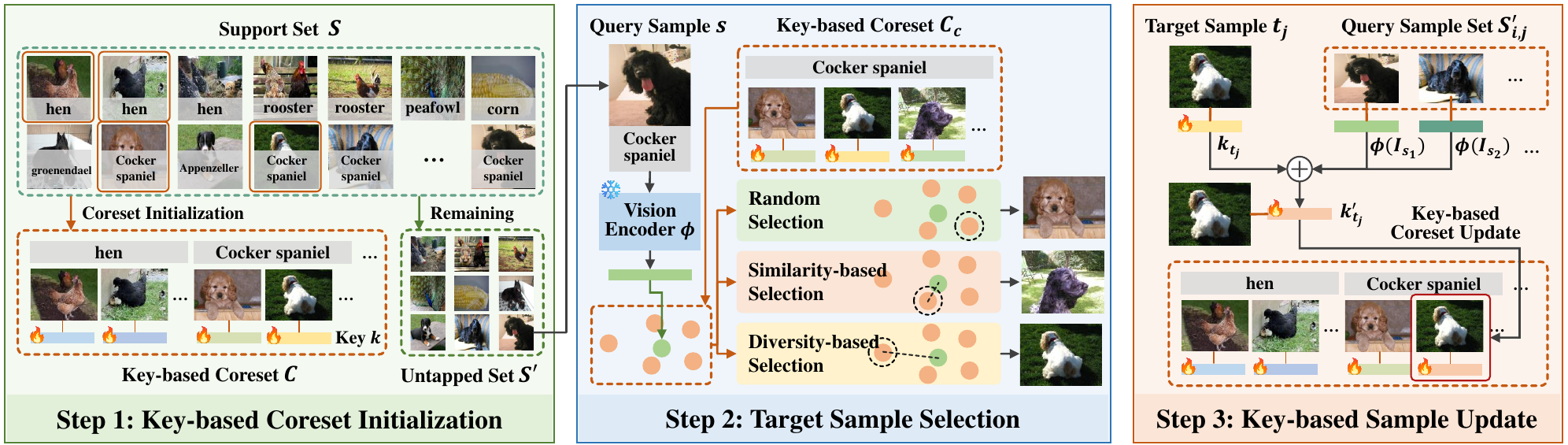}
  \caption{Overview of the KeCO framework: First, a subset is randomly selected from the support set $S$ to initialize the Coreset $C$, with the remaining samples forming the untapped set $S'$. A vision encoder is then used to extract visual features for all samples in $C$, which serve as their \textit{keys}. Next, for each query sample $s$ in the untapped set, its extracted feature will be used to select a corresponding target sample $t \in C$ based on three different strategies (Random Selection, Similarity-based Selection, Diversity Selection). Finally, the key of each $t$ in $C$ will be updated by aggregating the features of all associated query samples.}
  \label{fig:keco}
\end{figure*}

\section{Method}
\subsection{Preliminaries} 

The conventional In-context learning (ICL) within Large Vision-Language Model (LVLMs) is to construct an in-context sequence $\mathcal{D}$ by retrieving $n$ samples from a given support set. Specifically, given a support set $\mathcal{S} = \{(\bm{I}_i, \bm{y}_i)\}_{i=1}^{N}$, where $\bm{I}_i$ represents an image and $\bm{y}_i$ is its corresponding label, the in-context sequence is formed as follows:
\begin{equation} 
    \mathcal{D} = \{(\bm{I}_1, \bm{y}_1), (\bm{I}_2, \bm{y}_2), \dots, (\bm{I}_n, \bm{y}_n), (\hat{\bm{I}})\}.
\end{equation}
Here, the first $n$ image-label pairs serve as demonstrations, while $\hat{\bm{I}}$ is the test image whose label $\hat{y}$ needs to be predicted. The constructed sequence $\mathcal{D}$ is then fed into the LVLM, which generates the predicted label $\hat{y}$ based on the following probability distribution:
\begin{equation}
\hat{y} = \operatorname*{argmax}_{y \in Y} P(y \mid \mathcal{D}),
    \label{eq:prob}
\end{equation}
where $Y$ is the label space.

The selection of in-context examples $\{(\bm{I}_1, \bm{y}_1), \dots, (\bm{I}_n, \bm{y}_n)\}$ plays a crucial role in the performance of ICL, as it directly affects the contextual understanding and generalization capability of LVLMs. A common approach is similarity-based retrieval, where samples most similar to the test image $\hat{\bm{I}}$ are selected based on a feature extractor $\phi(\cdot)$, such as CLIP:
\begin{equation} 
    \text{sim}(\hat{\bm{I}}, \bm{I}_i) = \phi(\hat{\bm{I}})^\top \phi(\bm{I}_i), \quad \text{where } \bm{I}_i \in \mathcal{D}.
\end{equation}

However, this method becomes computationally expensive for a large support set. To improve efficiency, coreset-based methods first select a compact and representative subset $\mathcal{C} \subset \mathcal{S}$, then retrieve samples from $\mathcal{C}$ instead of the whole support set $\mathcal{S}$. The coreset can be constructed by samples with high diversity or large information gain, ensuring a more representative selection for ICL. However, due to the nature of the feature spaces in NLP and vision domain are fundamentally different,  coreset methods in the NLP domain still struggle to successfully apply to the vision domain.

\subsection{Key-based Coreset Optimization (KeCO)}
To address the aforementioned issues, we propose a straightforward but effective framework to obtain a compact and effective coreset, termed Key-based Coreset Optimization (KeCO), as shown in Figure~\ref{fig:keco}. Given a support set $\mathcal{S}$ of size $n$, we select a initial subset, referred to as the \textbf{Key-based Coreset}, denoted by $\mathcal{C}$, with a size of $m$, where $m<n$. The untapped samples in $\mathcal{S}$ form the untapped set $\mathcal{S'} = \mathcal{S} - \mathcal{C}$. $\mathcal{S'}$ will be used to update $\mathcal{C}$, and the updated coreset is denoted as $\mathcal{C'}$. As shown in Figure~\ref{fig:keco}, this framework consists of three stages: 
\begin{itemize}[leftmargin=*]
\item \textbf{Key-based Coreset Initialization}: A coreset $\mathcal{C}$ is first selected from the entire support set $\mathcal{S}$. The visual feature of the samples in $\mathcal{C}$ are extracted using a vision encoder, which serves as their corresponding keys for retrieval and updates. 
\item \textbf{Target Sample Selection}: Given a query sample $s$ from untapped set $S'$, a corresponding target sample $t$ from $\mathcal{C}$ will be selected using different selection strategies. 
\item \textbf{Key-based Sample Update}: An update strategy is applied to incorporate the information from the untapped query samples into each $t$, resulting in more informative coreset $\mathcal{C'}$.
\end{itemize}
In the following subsections, we detail each stage of KeCO.

\subsection{Key-based Coreset Initialization}
We adopt a simple \textbf{Random initialization} strategy to get the initial coreset. Specifically, given a support set $\mathcal{S} = \{(\bm{I}_i, \bm{y}_i)\}_{i=1}^{n}$ with $j$ categories, we construct $\mathcal{C}$ by randomly selecting $m$ samples while ensuring class balance. That is, for each category $c \in \{1, \dots, j\}$, we select  
\begin{equation} 
    \mathcal{C} = \bigcup_{c=1}^{j} \mathcal{C}_c, \quad \text{where} \quad 
    \mathcal{C}_c = \{ (\bm{I}_i, \bm{y}_i) \mid \bm{y}_i = c, \, (\bm{I}_i, \bm{y}_i) \in \mathcal{S} \},
\end{equation}
such that $|\mathcal{C}_c| = m/j$, where $\mathcal{C}_c$ denotes the subset of $\mathcal{C}$ belonging to the category $c$. For each selected sample, we extract its visual feature $\bm{k}$ using a vision encoder $\phi$, which serves as its \textbf{key} and is stored within $\mathcal{C}$. These keys are later used for retrieval and updating in subsequent stages. Thus, each element in $\mathcal{C}$ is represented as:
\begin{equation}
    \mathcal{C} = \{ (\bm{I}_i, \bm{y}_i, \bm{k}_i) \mid \bm{k}_i = \phi(\bm{I}_i), \, i = 1, \dots, m \}.
\end{equation}

Beyond random initialization, we also explore two alternative initialization strategies: \textbf{K-center initialization} and \textbf{InfoScore initialization}. The K-center approach aims to maximize diversity by selecting samples that best represent the overall distribution of $\mathcal{S}$, while the InfoScore method prioritizes samples with the highest information gain based on LVLM's feedback. Implementation details of these strategies are provided in Appendix A.1.

\subsection{Target Sample Selection}
\label{sec:target-sample-selection}
After Initialization, we use the untapped set $\mathcal{S'}$ to update $\mathcal{C}$. For each query sample $s = (\bm{I}_s, c)\in S'$, we need to determine which sample in $\mathcal{C}$ could be updated by $s$. We refer to this sample in $\mathcal{C}$ as the \textbf{target sample} \(t\). To effectively select the best $t$ for $s$, we first retrieve all samples in $C$ with the same class label $c$, denoted as $C_c$, then explore the following selection strategies to choose the corresponding target sample $t$ from $C_c$:

\begin{itemize}[leftmargin=*]
    \item \textbf{Random Selection (RS)}: The simplest approach is to randomly select a target sample from \(C_c\). While easy to implement, this method does not prioritize samples that may benefit most from updates.
    
    \item \textbf{Similarity-based Selection (SS)}: In this strategy, we select the sample that is most similar to \(s\) from \(C_c\). To measure similarity, we compute the cosine similarity between feature representations:
    \begin{equation}
    t = \arg\max_{e \in C_c} \frac{k_{e} \cdot \phi(\bm{I}_s)}{ \|k_{e}\|\|\phi(\bm{I}_s)\| },
\end{equation}

    where \(k_e\) denotes the key of the sample $e$ in $C$. This method ensures that the keys in $C$ are updated only by the most similar samples from the newly incoming data.

    \item \textbf{Diversity-based Selection (DS)}:  In this strategy, we select the sample that is least similar to \(s\). While this may initially appear counterintuitive, it has been shown to be the most effective strategy as demonstrated and discussed in our Section~\ref{sec4.4}:
    \begin{equation}
        t = \arg\min_{e \in C_c} \frac{k_{e} \cdot \phi(\bm{I}_s)}{ \|k_{e}\|\|\phi(\bm{I}_s)\| },
    \end{equation}
\end{itemize}

Once the target sample \(t\) is selected, it will undergo the key-based update process described in the next section.

\subsection{Key-based Sample Update}

We utilize untapped set $S'$ to update the keys in the coreset, dividing it into mini-batches for smooth updates:
\begin{equation}
    S' = \bigcup_{i=1}^{(n-m)/b} S'_i,
\end{equation}
where each batch $S'_i = \{s_{i_1}, s_{i_2}, \ldots, s_{i_b}\}$ contains $b$ samples.

For each sample $s \in S'_i$, we select a corresponding \textbf{target sample} $t \in \mathcal{C}$ from the coreset, based on the strategy introduced in Section~\ref{sec:target-sample-selection}. Since multiple samples in $S'_i$ may select the same target $t_j$, we group all such samples together:
\begin{equation}
    S'_{i,j} = \{s \in S'_i \mid \text{target}(s) = t_j\},
\end{equation}
where $\text{target}(s)$ denotes the selected target sample for $s$. This defines $S'_{i,j}$ as the subset of samples in batch $S'_i$ that share $t_j$ as their target.

Then, we update the key $k_{t_{j}}$ of $t_j$ using:
\begin{equation}
    k'_{t_{j}} = k_{t_{j}} - \alpha \cdot \frac{1}{|S'_{i,j}|} \sum_{s \in S'_{i,j}} (k_{t_{j}} - \phi(\bm{I}_s)),
\label{eq:update}
\end{equation}

where $|S'_{i,j}|$ is the number of samples associated with $t_j$, and $\alpha$ is a update rate that controls the size of the update step, which ranges from 0 to 1. This update strategy encourages each key to move toward the averaged features of its associated samples in a controlled manner, and the update process is repeated for $e$ epochs.

\subsection{Simulated Online Scenario}

In addition to the general coreset update using a fixed support set, we also explore a scenario that is more aligned with realistic and practical considerations: the \textbf{simulated online scenario}, where the untapped samples arrive in a streaming pattern. Specifically, we consider a data stream $\mathcal{S}^{\text{stream}} = \{(\bm{I}_1, \bm{y}_1), (\bm{I}_2, \bm{y}_2), \dots, (\bm{I}_n, \bm{y}_n)\}$ of size $n$, where each sample arrives sequentially. Our framework adapts naturally to this online scenario with minor modifications to the three core steps: key-based coreset initialization, target sample selection, and key-based sample update.

\textbf{Online Key-based Corese Initialization.} 
Unlike the general scenario where the coreset $\mathcal{C}$ is initialized all at once, the online scenario requires a \textbf{Filling-based initialization} due to the sequential data stream. To maintain class balance, we allocate a quota of $m/j$ samples per class for a total coreset size $m$ over $j$ classes. 
For each incoming sample $(\bm{I}_s, \bm{y}_s)$, if class $\bm{y}_s$ has fewer than $m/j$ samples in $\mathcal{C}$, it is directly added. Otherwise, it updates $\mathcal{C}$ via our key-based strategy.

\textbf{Online Target Sample Selection.}

To determine the target sample to update, we adopt the same selection strategies as in the general scenario: RS, SS, and DS, as described in Section~\ref{sec:target-sample-selection}.

\textbf{Online Key-based Sample Update.}
After selecting the target sample $t$ from the coreset, the incoming sample $(\bm{I}_s, \bm{y}_s)$ is used to update the key of $t$ using the same update rule introduced in Equation~(\ref{eq:update}). Since data arrives one sample at a time, the update is applied immediately upon the arrival of each new sample:
\begin{equation}
    k'_t = k_t - \alpha (k_t - \phi(\bm{I}_s)) = (1-\alpha)k_t + \alpha\cdot\phi(\bm{I}_s),
\label{eq:update_online}
\end{equation}
where $\phi(\bm{I}_s)$ is the visual representation of the incoming sample and $\alpha$ is the update rate that ranges from 0 to 1. This online update strategy enables the coreset to continually integrate new information without storing the entire data stream.

\subsection{Inference and Evaluation}

Once we obtain the final coreset $\mathcal{C'}$ by updating $\mathcal{C}$ via KeCO, we can evaluate a test image $\hat{\bm{I}}$ by retrieving the top-$k$ most relevant samples from $\mathcal{C}'$, based on the similarity between $\phi(\hat{\bm{I}})$ and all stored keys $\bm{k}$. These retrieved samples constitute the in-context sequence $\mathcal{D}$, which is concatenated with the test image and fed into the frozen LVLM for prediction, as defined in Equation~(\ref{eq:prob}).

\section{Experiments}
\begin{table*}
	\setlength{\tabcolsep}{4pt}
	\centering
	\begin{tabular}{c c ccccc c ccccc c ccccc} 
		\toprule
		\textbf{Dataset} &  & \multicolumn{5}{c}{\textbf{CUB-200}}    
          &  & \multicolumn{5}{c}{\textbf{Stanford Dogs }}  
         &  & \multicolumn{5}{c}{\textbf{ImageNet-100 }}                              \\ 
		\cline{1-1}\cline{3-7}\cline{9-13} \cline{15-19}
		\multirow{2}{*}{\textbf{Method}} & \multirow{2}{*}{} & \multicolumn{2}{c}{OF-3B}   &  & \multicolumn{2}{c}{IDE-8B} &  & \multicolumn{2}{c}{OF-3B}   &  & \multicolumn{2}{c}{IDE-8B}  & & \multicolumn{2}{c}{OF-3B}   &  & \multicolumn{2}{c}{IDE-8B} \\
		&                   & $ 2 $-shot & $ 4 $-shot & &  $ 2 $-shot & $ 4 $-shot  &  & $ 2 $-shot & $ 4 $-shot  &  & $ 2 $-shot & $ 4 $-shot &  & $ 2 $-shot & $ 4 $-shot  &  & $ 2 $-shot & $ 4 $-shot   \\ 
		\cline{1-1}\cline{3-7}\cline{9-13}\cline{15-19}
		FS-IC                        &                   & $ 48.11 $  & $ 34.88 $   &  & $ 84.65  $  & $ 90.36 $   & &  49.80 & $ 50.22 $   & & $ 74.42 $  & $ 82.25 $ &  & $ 65.26 $  & 62.92 & & 90.52 &94.98 \\
		FS-IS                        &                   & $ 61.04 $  & $ 58.61 $   &  & $ 85.62  $  & $ 93.09 $   & &  61.42 & $ 62.55 $   & & $ 76.89 $  & $ 87.23 $ &  & $ 71.88 $  & 71.98 & & 90.48 &95.50 \\  \hline

         KeCO-SS                        &                   & $ 49.65 $  & $ 32.22 $   &  & $ 84.60  $  & $ 90.58 $   & &  56.60 & $ 51.54 $   & & $ 75.59 $  & $ 82.27 $ &  & $ 70.66 $  & 63.60 & & 90.78 &95.88 \\
        KeCO-RS                        &                   & $ 72.90 $  & $ 72.66 $   &  & $ 85.45  $  & $ 94.34 $   & &  72.20 & $ 73.04 $   & & $ 79.66 $  & $ 90.45 $ &  & $ 77.50 $  & 79.30 & & 91.14 &96.12 \\
        KeCO-DS                        &                   & $ \textbf{74.20} $  & $ \textbf{76.99} $   &  & $ \textbf{87.38}  $  & $ \textbf{94.75} $   & &  \textbf{73.75} & $ \textbf{75.13} $   & & $ \textbf{79.72} $  & $ \textbf{91.72} $ &  & $ \textbf{78.22} $  & \textbf{80.28} & & \textbf{91.68} &\textbf{96.22} \\
        
        \hline

\textit{Online  Scenario}                       &                   &   &  &  &   &   & &  &    & &   &   &  &   &  & &  & \\ \hline        
          KeCO-SS                        &                   & $ 46.41 $  & $ 30.88 $   &  & $ 83.51  $  & $ 88.70 $   & &  54.63 & $ 50.89 $   & & $ 73.39 $  & $ 82.33 $ &  & $ 67.48 $  & 62.46 & & 90.22 &94.76 \\
		KeCO-RS                        &                   & $ 65.90 $  & $ 58.73 $   &  & $ 85.69  $  & $ 93.41 $   & &  67.06 & $ 67.52 $   & & $ 78.38 $  & $ 88.45 $ &  & $ 74.94 $  & 75.26 & & 90.58 &95.78 \\
		KeCO-DS                        &                   & $ \textbf{70.61} $  & $ \textbf{67.81} $   &  & $ \textbf{85.91}  $  & $ \textbf{93.77} $   & &  \textbf{69.62} & $ \textbf{72.37} $   & & $ \textbf{78.87} $  & $ \textbf{90.06} $ &  & $ \textbf{76.96} $  & \textbf{77.64} & & \textbf{90.88} &\textbf{95.94} \\ \hline
         
		\hline
	\end{tabular}
	\caption{Performance accuracy (\%) of OF-3B and IDE-8B on CUB-200 (coreset size = 1,000, support set size = 5,000), Stanford Dogs (coreset size = 1,200, support set size = 6,000) and ImageNet-100 (coreset size = 1,000, support set size = 5,000). Results are evaluated under different shot conditions (2-shot and 4-shot) across three KeCO methods and compared against two baselines (FS-IC and FS-IS).}

	\label{tab: all}
\end{table*}

\subsection{Models}
\noindent \textbf{LVLMs }
Given the limited number of LVLMs supporting multi-modal ICL, we employ OpenFlamingo-3B (OF-3B)~\cite{alayrac2022flamingo} and IDEFICS-8B (IDE-8B)~\cite{laurenccon2023obelics} to evaluate the ICL performance.

\noindent \textbf{Vision Encoder}
We choose to use CLIP/ViT-L-14 (CLIP)~\cite{radford2021learning} and google/siglip-so400m-patch14-384 (Siglip)~\cite{zhai2023sigmoid} to extract visual feature of sample's image, serving as the key for each sample. They are the corresponding visual language models (VLMs) for OF-3B and IDE-8B, respectively. This is done to align the representation space of LVLMs.


\subsection{Baselines}
To evaluate the effectiveness of KeCO, we compare our method with two traditional ICL baselines.

\noindent \textbf{Fewshot in Coreset (FS-IC).}
We evaluate the coreset $C$ that is constructed solely through key-based initialization from the support set $S$, without any refinement using the untapped samples $S'$. During ICL inference, demonstrations are selected from this unrefined coreset $C$ to form ICEs for prediction.

\noindent \textbf{Fewshot in Support Set (FS-IS).}
To evaluate the performance on the entire support set $S$, demonstrations are selected directly from $S$ to form ICEs for ICL inference, with no updates applied to the data.

\subsection{Implementation Details}
We utilize three image classification datasets, which include both coarse-grained and fine-grained categories. Specifically, they are CUB-200~\cite{welinder2010caltech}, Stanford Dogs~\cite{khosla2011novel}, and ImageNet-100~\cite{russakovsky2015imagenet}(details of these datasets are provided in Appendix A.2). We evaluate the in-context image classification capability of OF-3B, IDE-8B and Qwen2-VL~\cite{wang2024qwen2} in different methods. For OF-3B, due to its limited context length, we employ probabilistic inference for it. Specifically, for each class name, we compute the probability of the class name given the image and the prompt, denoted as prob(class name|image, ICE sequences), and select the class name with the highest probability as the prediction. Since class names can consist of multiple tokens (e.g., "tiger shark" consists of two tokens), we average the probabilities of all tokens~\cite{brown2020language}. For IDE-8B and Qwen2-VL~\cite{wang2024qwen2}, we provide a list of candidate choices in the prompt, and we follow~\cite{geigle2024african} to formulate image classification as a multiple-choice problem. In addition to the correct label, three other options are randomly selected from the classes excluding the correct label. Every in-context example is: "<image> Which of these choices is shown in the image? Choices: A.<class name A>, B.<class name B>, C. <class name C>, D. <class name D> Answer with the letter from the given choices directly."

We use top-1 accuracy as the metric to evaluate the performance due to its clarity and common usage. In general, each category in the coreset $C$ contains 5 or 10 samples (depending on the dataset), and the size $n$ of the support set $S$ is five times the size $m$ of $C$. Therefore, for CUB-200, $n=5000$ and $m=1000$; for Stanford Dogs, $n=6000$ and $m=1200$; and for ImageNet-100, $n=5000$ and $m=1000$. Since $S'=S-C$, the sizes of $S'$ are respectively 4,000, 4,800 and 4,000. Unless otherwise specified, the update rate $\alpha$ is set to 0.2, the epoch $e$ is set to 10, and the batch size $b$ is set to 1,000 (ablation study on $\alpha$, $e$ and $b$ are provided in Appendix A.3).

\subsection{Results and Key Findings.}
\label{sec4.4}
\begin{figure}
\centering\includegraphics[width=0.48\textwidth]{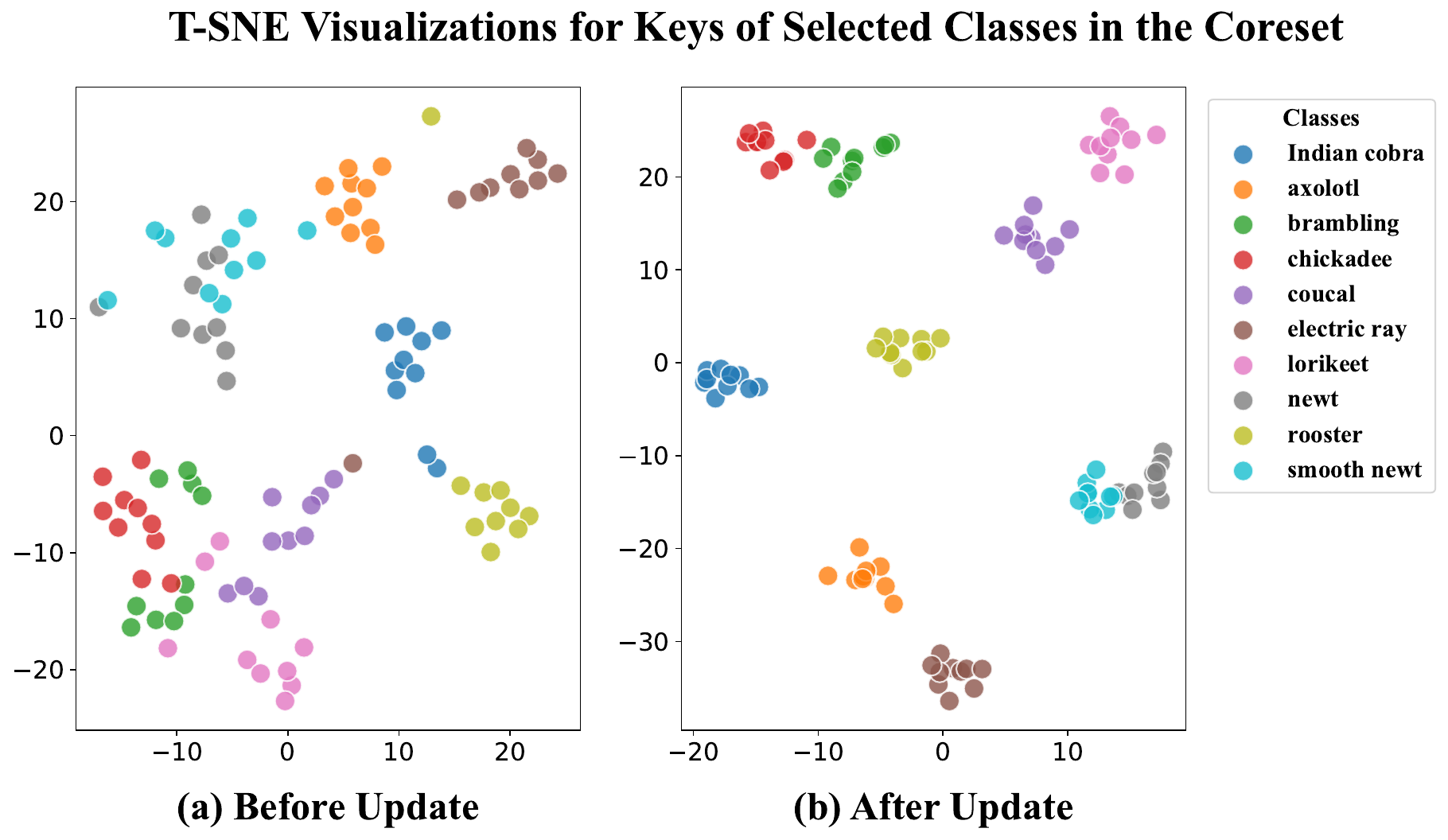}
  \caption{T-SNE visualization of the samples' keys of selected class in (a) the pre-update coreset and (b) the updated coreset using the KeCO-DS method.}
  \label{fig:tsne}
\end{figure}

\begin{figure*}
  \centering\includegraphics[width=0.9\textwidth]{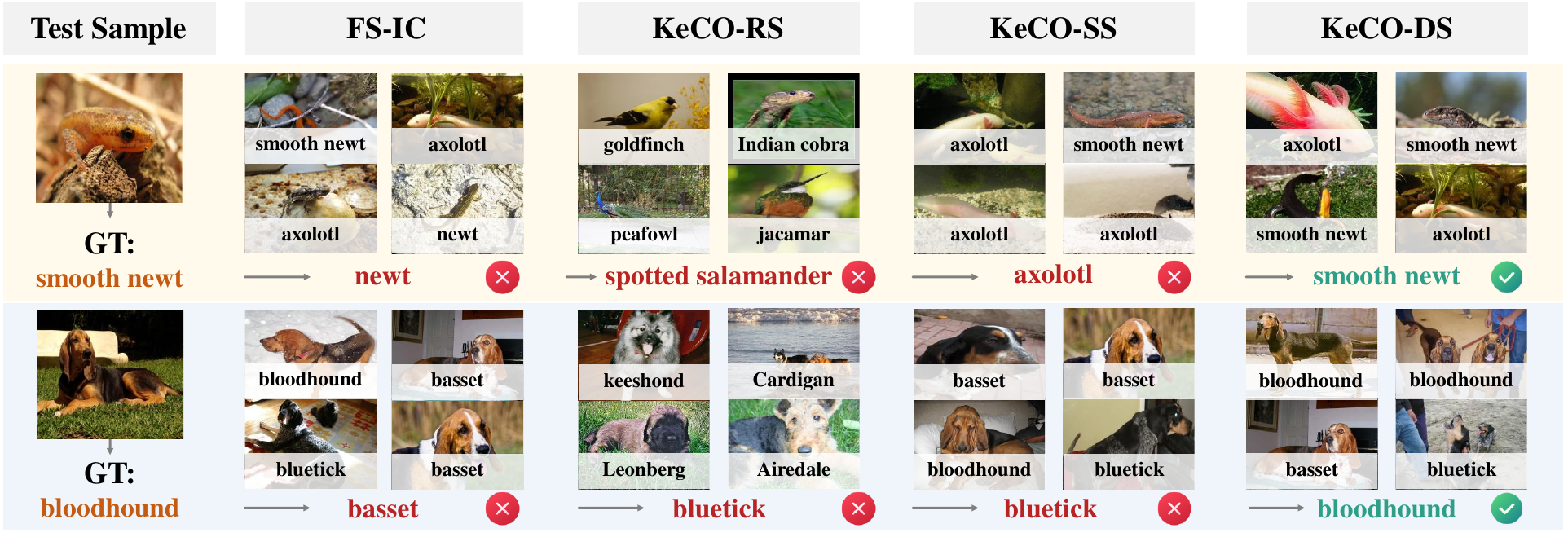}
  \caption{Case study on comparing FS-IC with three KeCO methods (RS, SS and DS) in selecting demonstrations and predicting for in-context learning.}
  \label{fig:case}
\end{figure*}

\noindent \textbf{Incorporating Information from Untapped Data into the Coreset Improves LVLM's ICL Performance. } 
Table~\ref{tab: all} clearly shows the ICL performance comparisons across three datasets and Figure~\ref{fig:case} illustrates two cases comparing FS-IC with three KeCO methods (RS, SS, and DS) in selecting demonstrations and making predictions for ICL. The results in Table~\ref{tab: all} indicate that OF-3B and IDE-8B, when using most KeCO methods (except KeCO-SS), consistently outperform FS-IC, despite identical coreset sizes. The primary  difference lies in the KeCO methods' utilization of keys as carriers for the information of untapped data. For instance, in a 4-shot setting compared to FS-IC, KeCO-RS and KeCO-DS enable OF-3B to achieve improvements of 37.78\% and 42.11\% in the CUB-200, 22.82\% and 24.91\% in the Stanford Dogs, and 16.38\% and 17.36\% in the ImageNet-100. Even for the already robust IDE-8B, these methods lead to an increase of 3.98\% and 4.39\% in the CUB-200, 8.2\% and 9.47\% in the Stanford Dogs, and 1.14\% and 1.24\% in the ImageNet-100.

When the coreset keys are updated with untapped data, each update integrates new information into coreset, enhancing the category-relevant information of each sample's key. As illustrated in Figure~\ref{fig:tsne}, in the original coreset, the keys or visual features of samples within the same category are quite dispersed. However, after updates with KeCO-DS, these keys become more clustered, making the distinction between different categories more pronounced. This clustering effect facilitates the retrieval of samples belonging to the same category as the test sample, thereby providing more precise and relevant knowledge to the LVLM when conducting ICL. 

While a larger support set can accommodate more information, as shown by FS-IS outperforming FS-IC, KeCO-RS and KeCO-DS show superior performance over FS-IS across three datasets, even though the support set size of FS-IS is five times larger than the coreset size of the KeCO methods. For instance, in a 2-shot setting, the ICL performance of OF-3B using the KeCO-DS achieves 74.02\%, compared to 61.04\% using FS-IS. Similarly, IDE-8B improves from 85.62\% with FS-IS to 87.38\% with the KeCO-DS on the CUB-200. This further validates the effectiveness of KeCO in enhancing the ICL performance of LVLMs.

The KeCO framework also offers a novel perspective for coreset selection in resource-constrained ICL scenarios, highlighting the potential benefits of optimizing the coreset with untapped data. While current research on retrieval of in-context examples primarily focuses on selecting representative subsets without updates or set the full training dataset as the support set, our findings suggest that even a simple randomly initialized subset can be an effective coreset after updated using untapped data.


\begin{figure}
  \centering\includegraphics[width=0.48\textwidth]{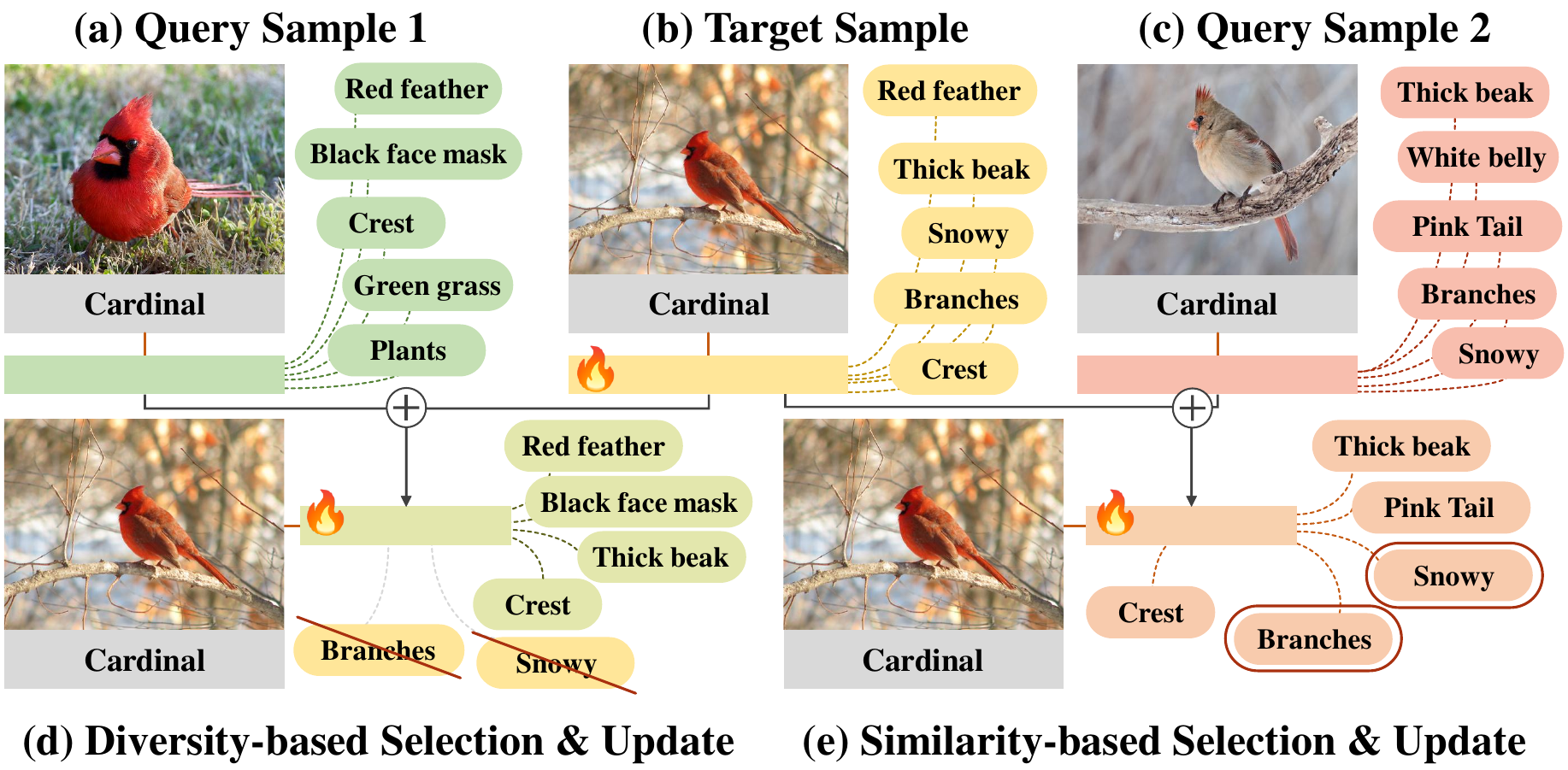}
  \caption{When blending keys from dissimilar samples in (d) Diversity-based selection, the updated key are encouraged to preserves shared, category-relevant information, while suppressing category-irrelevant or misleading attributes. In contrast, when blending similar samples in (e), all information, including misleading information, is retained in the updated key.}
  \label{fig:mixup}
\end{figure}

\noindent \textbf{Target Sample Selection is Important for KeCO. } 
From the experimental results, it is clear that the selection of target samples plays a crucial role in the effectiveness of KeCO. Both KeCO-RS and KeCO-DS lead to performance improvements, with KeCO-DS exhibiting the best results.  As we can see from Equation~\ref{eq:update_online}, $\alpha$ is a parameter that ranges from 0 to 1, which determines that $k'_{t}$ will lie somewhere between $k_{t}$ and $\phi(\bm{I}_s)$. Therefore, when a sample least similar to the query sample in untapped set $S'$ is selected for update, both keys are averaged with a weight of $\alpha$. This effectively maintains the invariant class features while blurring the attribute differences. As illustrated in Figure~\ref{fig:mixup} (d), the invariant part between the keys of the target (b) and query samples (a), such as `red feather' or `crest' related to the `Cardinal' category, are category-relevant information. They exist in both target and query samples, so they will still be preserved in the updated key after merge dissimilar samples in this class. However, category-irrelevant attributes, such as background information (`green grass' vs. `snowy'), are blended during the update process, because they only appear in one image. This blurs unnecessary distinctions and enables KeCO-DS to focus more effectively on category-relevant features essential for classification.


On the other hand, KeCO-SS performs the worst, sometimes even underperforming FS-IC. This occurs when the query samples in $S'$ select the most similar sample for update, potentially retaining not only useful but also category-irrelevant information, such as background information, in the updated key. As depicted in Figure~\ref{fig:mixup} (e), this is the case when the category-irrelevant information (`snowy' and `branches') is also preserved in the updated key. Consequently, samples with the same category-irrelevant information but from different categories are more likely to be retrieved as ICE when retrieving from the coreset. This could mislead the LVLMs, negatively affecting their ICL performance.

\noindent \textbf{KeCO Methods Perform Better on Fine-grained Classification Dataset than on Coarse-grained Dataset. }
In fine-grained classification datasets, the similarity between categories is higher compared to coarse-grained classification, making it more challenging to distinguish between samples of different sub-classes. Table~\ref{tab: all} shows that KeCO-RS and KeCO-DS improvements over the baselines are more pronounced on fine-grained classification datasets (CUB-200 and Stanford Dogs) compared to the coarse-grained dataset (ImageNet-100). For instance, for OF-3B, the 2-shot ICL on CUB-200 improves from 61.04\% to 74.20\%, an increase of 13.16\%, whereas on ImageNet-100, it improves from 71.88\% to 78.22\%, an increase of 6.34\%. This discrepancy is due to the fact that the pre-training corpus of LVLMs is unbalanced, with a majority being coarse-grained knowledge~\cite{li2023m,liu2024survey}. Therefore, LVLMs inherently possess sufficient knowledge about coarse-grained categories but lack knowledge of subordinate-level categories. This can be evidenced from Table~\ref{tab: all} where OF-3B and IDE-8B's FS-IC and FS-IS performance on CUB-200 and Stanford Dogs are lower than those on ImageNet-100.

After we optimzie the keys in the coreset, we can retrieve better demonstrations, which supplements useful knowledge when LVLMs perform ICL inference, thereby having a larger impact on fine-grained datasets. However, despite the larger improvement on fine-grained datasets compared to ImageNet, the in-context classification accuracy still does not surpass that on ImageNet, suggesting a need for more balanced pre-training corpora for training LVLMs.


\noindent \textbf{KeCO Methods Retain a Strong Performance in a Simulated Online Scenario. }
From Table~\ref{tab: all}, it can be observed that the ICL performance of LVLMs using KeCO-RS and KeCO-DS consistently outperforms two baselines (FS-IC and FS-IS) in a simulated online scenario.  For instance, when using KeCO-DS, in the 4-shot setting, the ICL performance of OF-3B on the Stanford Dogs is about 22.15\% and 9.82\% higher than FS-IC and FS-IS, respectively. Similarly, for IDE-8B, there is an improvement of 7.81\% and 2.83\%, respectively. In reality, systems usually receive a continuous stream of data, rather than having access to large amounts of data at once that can be reused. Therefore, the improved performance of LVLMs using KeCO-RS and KeCO-DS in the simulated online scenario holds more practical significance, as it is more reflective of real-world situations.

Furthermore, we performed an additional experiment on the high-performance commercial model Qwen2-VL~\cite{wang2024qwen2}, applying KeCO-DS to update coreset in an online scenario. This approach aligns with real-world applications where data typically arrives in a stream, necessitating models that can continuously adapt. As shown in Table~\ref{tab:qwen}, in the online scenario, the 2-shot ICL performance on the CUB-200 improved from 92.95\% to 98.30\%. On the Stanford Dogs, the performance improved by 1\%, from an already high baseline of 96.92\%, after updating the coreset. This highlights KeCO's efficacy in enhancing the adaptability and responsiveness of high-performance LVLMs like Qwen2-VL to online data with minimal computational overhead.
\definecolor{Gray}{rgb}{0.501,0.501,0.501}
\definecolor{GuardsmanRed}{rgb}{0.815,0,0}
\definecolor{MilanoRed}{rgb}{0.717,0.05,0.05}
\begin{table}
\centering
\small
\setlength{\tabcolsep}{1pt}
\begin{tblr}{
  cells = {c},
  column{2} = {fg=Gray},
  column{3,5} = {fg=black},
  column{5} = {fg=GuardsmanRed},
  hline{1,4} = {-}{0.08em},
  hline{2} = {-}{},
}
\textbf{Dataset} & \textbf{FS-IC} & \textbf{FS-IS} & \textbf{Online} &  $\Delta$\\
CUB-200          & 92.95    & 93.11      & 98.30              &  5.35 \\
Stanford Dogs    & 96.92    & 97.78      & 97.93               &   1.01\\
\end{tblr}
\caption{2-shot ICL Performance accuracy (\%) of Qwen2-VL on CUB-200 and Stanford Dogs. Results are evaluated under KeCO-DS in online scenario and two baselines (FS-IC and FS-IS). }
\vspace{-0.1cm}
\label{tab:qwen}
\end{table}

\section{Further Analyses}

\begin{table}
	\setlength{\tabcolsep}{2pt}
	\centering
	\begin{tabular}{c c ccc c ccc} 
		\toprule
		\multirow{2}{*}{\textbf{Method}} &\multirow{2}{*}{}& \multicolumn{3}{c}{OF-3B}   &  & \multicolumn{3}{c}{IDE-8B}  \\
		&                   & k-center & info & random & &  k-center & info  & random \\ 
		\cline{1-1}\cline{3-5}\cline{7-9}
		FS-IC    &  & 39.88   & 47.56 & 50.22 &   & 73.82 &81.38 & 82.25 \\
		FS-IS    &  & 62.55 & 62.55   & 62.55 &   & 87.23 &87.23 & 87.23   \\ 
         KeCO-RS     &  & 73.26  & 73.10  & 73.04  &    & 90.64 & 90.54  & 90.45  \\
        KeCO-DS     &  & 75.06   & 75.22 & 75.13  &    & 91.52 & 91.38 & 91.72  \\
        \hline

\textit{Online  Scenario}                 &   &  &  &   &  \\ \hline        
		KeCO-RS  &   & 55.07  & 66.67 & 67.52 &   & 87.51 & 88.72 & 88.45  \\
		KeCO-DS  &   & 58.82  & 69.81 &  72.37  & & 89.38 & 89.64 & 90.06  \\ \hline

	\end{tabular}
	\caption{4-shot ICL performance accuracy (\%) of OF-3B and IDE-8B on Stanford Dogs under different coreset initialization. Results are evaluated across two KeCO methods (KeCO-RS and KeCO-DS) and two baselines (FS-IC and FS-IS).}
	\label{tab: initial}
\end{table}

\subsection{Analyses of Different Coreset Initialization}
The K-center-greedy (k-center) approach aims to maximize diversity by selecting samples that best represent the overall distribution of the support set. The Infoscore approach prioritizes samples with the highest information gain based on LVLM. After initializing the coreset with these approaches, we used KeCO-RS and KeCO-DS for optimization. As shown in Table~\ref{tab: initial}, the ICL performance of two LVLMs with coresets initialized using the k-center is inferior to those initialized using Infoscore and random methods. This suggests that the k-center-greedy algorithm, despite its popularity in active learning, might not suit ICL. The primary goal in active learning is to diversify the labeled data to enhance the robustness of the trained model~\cite{ren2021survey}. However, in the context of ICL, the model relies on the in-context examples (ICE) to quickly adapt to downstream tasks. Therefore, it is crucial for these ICE to be closely related to the test input's label to furnish the model with pertinent knowledge. If the data in the coreset is overly diverse, it may encompass numerous samples with images filled with irrelevant and misleading information, such as images with overly conspicuous background. This could potentially interfere with the LVLM's inference process and diminish its ICL performance.

Furthermore, the ICL performance of LVLMs with a coreset initialized using Infoscore is inferior to that with a coreset initialized randomly. This can be attributed to the inadequate alignment of vision encoder and LLM in LVLMs, typically caused by the LLM's larger scale and the scarcity of high-quality multimodal datasets~\cite{he2025analyzing}. As a result, even the seemingly `useful' samples selected via Infoscore may not function effectively in Multimodal ICL. Moreover, studies such as~\cite{baldassini2024makes,zhang2024visually} have demonstrated that Multimodal ICL primarily focuses on text, overshadowing the role played by images. Therefore, the Infoscore obtained by LVLMs is likely to rely more on text information rather than visual information, leading to a deficiency in the evaluation of samples in terms of visual information. Consequently, the Infoscore metric, while applicable in NLP tasks, may not be as effective in more complex multimodal scenarios.

Despite the varying effectiveness of coresets initialized with different methods in the FS-IC setting, the use of the KeCO framework to update coreset leads to substantial improvements, even in the online scenario. For example, with coresets initialized using the k-center-greedy method, OF-3B sees improvements and 35.18\% in the common setting and 18.94\% in the online scenario after applying KeCO-DS, compared to the results with FS-IC. Similarly, IDE-8B experiences improvements of 15.56\% and 17.7\% in the common and online settings, respectively. These results further underscore the efficacy of the KeCO framework.

\subsection{Analyses of the Size of Additional Data and Coreset}
From Table 6 in Appendix A.4, it can be observed that as the amount of untapped data increases, there is an improvement in the FS-IS, KeCO-RS, and KeCO-DS settings. For instance, when the ratio of coreset size to untapped data size increases from 1:2 to 1:6, the performance of OF-3B under the FS-IS setting improves from 58.67\% to 63.42\%, and IDE-8B under the same setting improves from 75.68\% to 77.45\%. Similarly, under the KeCO framework, when the ratio changes from 1:2 to 1:6, OF-3B's ICL performance under the KeCO-DS setting improves from 69.99\% to 74.58\%, and IDE-8B under the same setting improves from 79.02\% to 79.92\%. The analysis of the coreset size can be found in the Appendix A.4.

\section{Conclusion}
In our paper, we propose a novel coreset optimization framework, KeCO, which effectively constructs a compact and effective coreset from a large support set. KeCO everages untapped data to aggregate category-relevent information into the coreset via feature-level updates. KeCO achieves larger gains on fine-grained datasets, highlighting its effectiveness in capturing subtle visual distinctions. In addition, our results indicate that diversity-based selection consistently outperforms other strategies, shedding light on better update mechanisms. Our experiments also demonstrate the effectiveness of KeCO in both conventional ICL settings and the proposed simulated online scenario, which better reflects practical use cases.

\begin{acks}
This work is supported by the National Science Foundation of China (62206048), the Natural Science Foundation of Jiangsu Province (BK20220819), and the Fundamental Research Funds for the Central Universities (2242025K30024). This research work is also supported by the Big Data Computing Center of Southeast University
\end{acks}


\bibliographystyle{ACM-Reference-Format}
\balance
\bibliography{main}


\begin{thebibliography}{52}


\ifx \showCODEN    \undefined \def \showCODEN     #1{\unskip}     \fi
\ifx \showISBNx    \undefined \def \showISBNx     #1{\unskip}     \fi
\ifx \showISBNxiii \undefined \def \showISBNxiii  #1{\unskip}     \fi
\ifx \showISSN     \undefined \def \showISSN      #1{\unskip}     \fi
\ifx \showLCCN     \undefined \def \showLCCN      #1{\unskip}     \fi
\ifx \shownote     \undefined \def \shownote      #1{#1}          \fi
\ifx \showarticletitle \undefined \def \showarticletitle #1{#1}   \fi
\ifx \showURL      \undefined \def \showURL       {\relax}        \fi
\providecommand\bibfield[2]{#2}
\providecommand\bibinfo[2]{#2}
\providecommand\natexlab[1]{#1}
\providecommand\showeprint[2][]{arXiv:#2}

\bibitem[Alayrac et~al\mbox{.}(2022)]%
        {alayrac2022flamingo}
\bibfield{author}{\bibinfo{person}{Jean-Baptiste Alayrac}, \bibinfo{person}{Jeff Donahue}, \bibinfo{person}{Pauline Luc}, \bibinfo{person}{Antoine Miech}, \bibinfo{person}{Iain Barr}, \bibinfo{person}{Yana Hasson}, \bibinfo{person}{Karel Lenc}, \bibinfo{person}{Arthur Mensch}, \bibinfo{person}{Katherine Millican}, \bibinfo{person}{Malcolm Reynolds}, {et~al\mbox{.}}} \bibinfo{year}{2022}\natexlab{}.
\newblock \showarticletitle{Flamingo: a visual language model for few-shot learning}.
\newblock \bibinfo{journal}{\emph{Advances in neural information processing systems}}  \bibinfo{volume}{35} (\bibinfo{year}{2022}), \bibinfo{pages}{23716--23736}.
\newblock


\bibitem[Awadalla et~al\mbox{.}(2023)]%
        {awadalla2023openflamingo}
\bibfield{author}{\bibinfo{person}{Anas Awadalla}, \bibinfo{person}{Irena Gao}, \bibinfo{person}{Josh Gardner}, \bibinfo{person}{Jack Hessel}, \bibinfo{person}{Yusuf Hanafy}, \bibinfo{person}{Wanrong Zhu}, \bibinfo{person}{Kalyani Marathe}, \bibinfo{person}{Yonatan Bitton}, \bibinfo{person}{Samir Gadre}, \bibinfo{person}{Shiori Sagawa}, {et~al\mbox{.}}} \bibinfo{year}{2023}\natexlab{}.
\newblock \showarticletitle{Openflamingo: An open-source framework for training large autoregressive vision-language models}.
\newblock \bibinfo{journal}{\emph{arXiv preprint arXiv:2308.01390}} (\bibinfo{year}{2023}).
\newblock


\bibitem[Baldassini et~al\mbox{.}(2024)]%
        {baldassini2024makes}
\bibfield{author}{\bibinfo{person}{Folco~Bertini Baldassini}, \bibinfo{person}{Mustafa Shukor}, \bibinfo{person}{Matthieu Cord}, \bibinfo{person}{Laure Soulier}, {and} \bibinfo{person}{Benjamin Piwowarski}.} \bibinfo{year}{2024}\natexlab{}.
\newblock \showarticletitle{What makes multimodal in-context learning work?}. In \bibinfo{booktitle}{\emph{Proceedings of the IEEE/CVF Conference on Computer Vision and Pattern Recognition}}. \bibinfo{pages}{1539--1550}.
\newblock


\bibitem[Brown et~al\mbox{.}(2020)]%
        {brown2020language}
\bibfield{author}{\bibinfo{person}{Tom Brown}, \bibinfo{person}{Benjamin Mann}, \bibinfo{person}{Nick Ryder}, \bibinfo{person}{Melanie Subbiah}, \bibinfo{person}{Jared~D Kaplan}, \bibinfo{person}{Prafulla Dhariwal}, \bibinfo{person}{Arvind Neelakantan}, \bibinfo{person}{Pranav Shyam}, \bibinfo{person}{Girish Sastry}, \bibinfo{person}{Amanda Askell}, {et~al\mbox{.}}} \bibinfo{year}{2020}\natexlab{}.
\newblock \showarticletitle{Language models are few-shot learners}.
\newblock \bibinfo{journal}{\emph{Advances in neural information processing systems}}  \bibinfo{volume}{33} (\bibinfo{year}{2020}), \bibinfo{pages}{1877--1901}.
\newblock


\bibitem[Chang and Jia(2022)]%
        {chang2022data}
\bibfield{author}{\bibinfo{person}{Ting-Yun Chang} {and} \bibinfo{person}{Robin Jia}.} \bibinfo{year}{2022}\natexlab{}.
\newblock \showarticletitle{Data curation alone can stabilize in-context learning}.
\newblock \bibinfo{journal}{\emph{arXiv preprint arXiv:2212.10378}} (\bibinfo{year}{2022}).
\newblock


\bibitem[Das et~al\mbox{.}(2024)]%
        {Das_2024_ECCV}
\bibfield{author}{\bibinfo{person}{Anurag Das}, \bibinfo{person}{Xinting Hu}, \bibinfo{person}{Li Jiang}, {and} \bibinfo{person}{Bernt Schiele}.} \bibinfo{year}{2024}\natexlab{}.
\newblock \showarticletitle{MTA-CLIP: Language-Guided Semantic Segmentation with Mask-Text Alignment}. In \bibinfo{booktitle}{\emph{Proceedings of the European Conference on Computer Vision (ECCV)}}.
\newblock


\bibitem[Dong et~al\mbox{.}(2022)]%
        {dong2022survey}
\bibfield{author}{\bibinfo{person}{Qingxiu Dong}, \bibinfo{person}{Lei Li}, \bibinfo{person}{Damai Dai}, \bibinfo{person}{Ce Zheng}, \bibinfo{person}{Jingyuan Ma}, \bibinfo{person}{Rui Li}, \bibinfo{person}{Heming Xia}, \bibinfo{person}{Jingjing Xu}, \bibinfo{person}{Zhiyong Wu}, \bibinfo{person}{Tianyu Liu}, {et~al\mbox{.}}} \bibinfo{year}{2022}\natexlab{}.
\newblock \showarticletitle{A survey on in-context learning}.
\newblock \bibinfo{journal}{\emph{arXiv preprint arXiv:2301.00234}} (\bibinfo{year}{2022}).
\newblock


\bibitem[Feng et~al\mbox{.}(2025)]%
        {feng2025redefining}
\bibfield{author}{\bibinfo{person}{Fu Feng}, \bibinfo{person}{Yucheng Xie}, \bibinfo{person}{Xu Yang}, \bibinfo{person}{Jing Wang}, {and} \bibinfo{person}{Xin Geng}.} \bibinfo{year}{2025}\natexlab{}.
\newblock \showarticletitle{Redefining< creative> in dictionary: Towards an enhanced semantic understanding of creative generation}. In \bibinfo{booktitle}{\emph{Proceedings of the Computer Vision and Pattern Recognition Conference}}. \bibinfo{pages}{18444--18454}.
\newblock


\bibitem[Geigle et~al\mbox{.}(2024)]%
        {geigle2024african}
\bibfield{author}{\bibinfo{person}{Gregor Geigle}, \bibinfo{person}{Radu Timofte}, {and} \bibinfo{person}{Goran Glava{\v{s}}}.} \bibinfo{year}{2024}\natexlab{}.
\newblock \showarticletitle{African or european swallow? benchmarking large vision-language models for fine-grained object classification}.
\newblock \bibinfo{journal}{\emph{arXiv preprint arXiv:2406.14496}} (\bibinfo{year}{2024}).
\newblock


\bibitem[He et~al\mbox{.}(2025)]%
        {he2025analyzing}
\bibfield{author}{\bibinfo{person}{Hulingxiao He}, \bibinfo{person}{Geng Li}, \bibinfo{person}{Zijun Geng}, \bibinfo{person}{Jinglin Xu}, {and} \bibinfo{person}{Yuxin Peng}.} \bibinfo{year}{2025}\natexlab{}.
\newblock \showarticletitle{Analyzing and Boosting the Power of Fine-Grained Visual Recognition for Multi-modal Large Language Models}.
\newblock \bibinfo{journal}{\emph{arXiv preprint arXiv:2501.15140}} (\bibinfo{year}{2025}).
\newblock


\bibitem[He et~al\mbox{.}(2020)]%
        {he2020momentum}
\bibfield{author}{\bibinfo{person}{Kaiming He}, \bibinfo{person}{Haoqi Fan}, \bibinfo{person}{Yuxin Wu}, \bibinfo{person}{Saining Xie}, {and} \bibinfo{person}{Ross Girshick}.} \bibinfo{year}{2020}\natexlab{}.
\newblock \showarticletitle{Momentum contrast for unsupervised visual representation learning}. In \bibinfo{booktitle}{\emph{Proceedings of the IEEE/CVF conference on computer vision and pattern recognition}}. \bibinfo{pages}{9729--9738}.
\newblock


\bibitem[Hu et~al\mbox{.}(2024)]%
        {Hu_2024_CVPR}
\bibfield{author}{\bibinfo{person}{Xinting Hu}, \bibinfo{person}{Li Jiang}, {and} \bibinfo{person}{Bernt Schiele}.} \bibinfo{year}{2024}\natexlab{}.
\newblock \showarticletitle{Training Vision Transformers for Semi-Supervised Semantic Segmentation}. In \bibinfo{booktitle}{\emph{Proceedings of the IEEE/CVF Conference on Computer Vision and Pattern Recognition (CVPR)}}.
\newblock


\bibitem[Huang et~al\mbox{.}(2023)]%
        {huang2023visual}
\bibfield{author}{\bibinfo{person}{Jiaxing Huang}, \bibinfo{person}{Jingyi Zhang}, \bibinfo{person}{Kai Jiang}, \bibinfo{person}{Han Qiu}, {and} \bibinfo{person}{Shijian Lu}.} \bibinfo{year}{2023}\natexlab{}.
\newblock \showarticletitle{Visual instruction tuning towards general-purpose multimodal model: A survey}.
\newblock \bibinfo{journal}{\emph{arXiv preprint arXiv:2312.16602}} (\bibinfo{year}{2023}).
\newblock


\bibitem[Jiang et~al\mbox{.}(2025)]%
        {Jiang_2025_CVPR}
\bibfield{author}{\bibinfo{person}{Yuchu Jiang}, \bibinfo{person}{Jiale Fu}, \bibinfo{person}{Chenduo Hao}, \bibinfo{person}{Xinting Hu}, \bibinfo{person}{Yingzhe Peng}, \bibinfo{person}{Xin Geng}, {and} \bibinfo{person}{Xu Yang}.} \bibinfo{year}{2025}\natexlab{}.
\newblock \showarticletitle{Mimic In-Context Learning for Multimodal Tasks}. In \bibinfo{booktitle}{\emph{Proceedings of the Computer Vision and Pattern Recognition Conference (CVPR)}}. \bibinfo{pages}{29825--29835}.
\newblock


\bibitem[Khosla et~al\mbox{.}(2011)]%
        {khosla2011novel}
\bibfield{author}{\bibinfo{person}{Aditya Khosla}, \bibinfo{person}{Nityananda Jayadevaprakash}, \bibinfo{person}{Bangpeng Yao}, {and} \bibinfo{person}{Fei-Fei Li}.} \bibinfo{year}{2011}\natexlab{}.
\newblock \showarticletitle{Novel dataset for fine-grained image categorization: Stanford dogs}. In \bibinfo{booktitle}{\emph{Proc. CVPR workshop on fine-grained visual categorization (FGVC)}}, Vol.~\bibinfo{volume}{2}.
\newblock


\bibitem[Lauren{\c{c}}on et~al\mbox{.}(2023)]%
        {laurenccon2023obelics}
\bibfield{author}{\bibinfo{person}{Hugo Lauren{\c{c}}on}, \bibinfo{person}{Lucile Saulnier}, \bibinfo{person}{L{\'e}o Tronchon}, \bibinfo{person}{Stas Bekman}, \bibinfo{person}{Amanpreet Singh}, \bibinfo{person}{Anton Lozhkov}, \bibinfo{person}{Thomas Wang}, \bibinfo{person}{Siddharth Karamcheti}, \bibinfo{person}{Alexander Rush}, \bibinfo{person}{Douwe Kiela}, {et~al\mbox{.}}} \bibinfo{year}{2023}\natexlab{}.
\newblock \showarticletitle{Obelics: An open web-scale filtered dataset of interleaved image-text documents}.
\newblock \bibinfo{journal}{\emph{Advances in Neural Information Processing Systems}}  \bibinfo{volume}{36} (\bibinfo{year}{2023}), \bibinfo{pages}{71683--71702}.
\newblock


\bibitem[Lauren{\c{c}}on et~al\mbox{.}(2024)]%
        {laurenccon2024obelics}
\bibfield{author}{\bibinfo{person}{Hugo Lauren{\c{c}}on}, \bibinfo{person}{Lucile Saulnier}, \bibinfo{person}{L{\'e}o Tronchon}, \bibinfo{person}{Stas Bekman}, \bibinfo{person}{Amanpreet Singh}, \bibinfo{person}{Anton Lozhkov}, \bibinfo{person}{Thomas Wang}, \bibinfo{person}{Siddharth Karamcheti}, \bibinfo{person}{Alexander Rush}, \bibinfo{person}{Douwe Kiela}, {et~al\mbox{.}}} \bibinfo{year}{2024}\natexlab{}.
\newblock \showarticletitle{Obelics: An open web-scale filtered dataset of interleaved image-text documents}.
\newblock \bibinfo{journal}{\emph{Advances in Neural Information Processing Systems}}  \bibinfo{volume}{36} (\bibinfo{year}{2024}).
\newblock


\bibitem[Levy et~al\mbox{.}(2023)]%
        {levy2023diverse}
\bibfield{author}{\bibinfo{person}{Itay Levy}, \bibinfo{person}{Ben Bogin}, {and} \bibinfo{person}{Jonathan Berant}.} \bibinfo{year}{2023}\natexlab{}.
\newblock \showarticletitle{Diverse Demonstrations Improve In-context Compositional Generalization}. In \bibinfo{booktitle}{\emph{Proceedings of the 61st Annual Meeting of the Association for Computational Linguistics (Volume 1: Long Papers)}}. \bibinfo{pages}{1401--1422}.
\newblock


\bibitem[Li et~al\mbox{.}(2023f)]%
        {li2023otterhd}
\bibfield{author}{\bibinfo{person}{Bo Li}, \bibinfo{person}{Peiyuan Zhang}, \bibinfo{person}{Jingkang Yang}, \bibinfo{person}{Yuanhan Zhang}, \bibinfo{person}{Fanyi Pu}, {and} \bibinfo{person}{Ziwei Liu}.} \bibinfo{year}{2023}\natexlab{f}.
\newblock \showarticletitle{Otterhd: A high-resolution multi-modality model}.
\newblock \bibinfo{journal}{\emph{arXiv preprint arXiv:2311.04219}} (\bibinfo{year}{2023}).
\newblock


\bibitem[Li et~al\mbox{.}(2023d)]%
        {li2023mimic}
\bibfield{author}{\bibinfo{person}{Bo Li}, \bibinfo{person}{Yuanhan Zhang}, \bibinfo{person}{Liangyu Chen}, \bibinfo{person}{Jinghao Wang}, \bibinfo{person}{Fanyi Pu}, \bibinfo{person}{Jingkang Yang}, \bibinfo{person}{Chunyuan Li}, {and} \bibinfo{person}{Ziwei Liu}.} \bibinfo{year}{2023}\natexlab{d}.
\newblock \showarticletitle{Mimic-it: Multi-modal in-context instruction tuning}.
\newblock \bibinfo{journal}{\emph{arXiv preprint arXiv:2306.05425}} (\bibinfo{year}{2023}).
\newblock


\bibitem[Li et~al\mbox{.}(2023e)]%
        {li2023ottermultimodalmodelincontext}
\bibfield{author}{\bibinfo{person}{Bo Li}, \bibinfo{person}{Yuanhan Zhang}, \bibinfo{person}{Liangyu Chen}, \bibinfo{person}{Jinghao Wang}, \bibinfo{person}{Jingkang Yang}, {and} \bibinfo{person}{Ziwei Liu}.} \bibinfo{year}{2023}\natexlab{e}.
\newblock \showarticletitle{Otter: A Multi-Modal Model with In-Context Instruction Tuning}.
\newblock \bibinfo{journal}{\emph{arXiv preprint arXiv:2305.03726}} (\bibinfo{year}{2023}).
\newblock


\bibitem[Li et~al\mbox{.}(2023a)]%
        {li2023blip}
\bibfield{author}{\bibinfo{person}{Junnan Li}, \bibinfo{person}{Dongxu Li}, \bibinfo{person}{Silvio Savarese}, {and} \bibinfo{person}{Steven Hoi}.} \bibinfo{year}{2023}\natexlab{a}.
\newblock \showarticletitle{Blip-2: Bootstrapping language-image pre-training with frozen image encoders and large language models}. In \bibinfo{booktitle}{\emph{International conference on machine learning}}. PMLR, \bibinfo{pages}{19730--19742}.
\newblock


\bibitem[Li et~al\mbox{.}(2024)]%
        {li2024configure}
\bibfield{author}{\bibinfo{person}{Li Li}, \bibinfo{person}{Jiawei Peng}, \bibinfo{person}{Huiyi Chen}, \bibinfo{person}{Chongyang Gao}, {and} \bibinfo{person}{Xu Yang}.} \bibinfo{year}{2024}\natexlab{}.
\newblock \showarticletitle{How to configure good in-context sequence for visual question answering}. In \bibinfo{booktitle}{\emph{Proceedings of the IEEE/CVF Conference on Computer Vision and Pattern Recognition}}. \bibinfo{pages}{26710--26720}.
\newblock


\bibitem[Li et~al\mbox{.}(2023c)]%
        {li2023m}
\bibfield{author}{\bibinfo{person}{Lei Li}, \bibinfo{person}{Yuwei Yin}, \bibinfo{person}{Shicheng Li}, \bibinfo{person}{Liang Chen}, \bibinfo{person}{Peiyi Wang}, \bibinfo{person}{Shuhuai Ren}, \bibinfo{person}{Mukai Li}, \bibinfo{person}{Yazheng Yang}, \bibinfo{person}{Jingjing Xu}, \bibinfo{person}{Xu Sun}, {et~al\mbox{.}}} \bibinfo{year}{2023}\natexlab{c}.
\newblock \showarticletitle{M\textsuperscript{3}IT: A Large-Scale Dataset towards Multi-Modal Multilingual Instruction Tuning}.
\newblock \bibinfo{journal}{\emph{arXiv preprint arXiv:2306.04387}} (\bibinfo{year}{2023}).
\newblock


\bibitem[Li et~al\mbox{.}(2023b)]%
        {li2023unified}
\bibfield{author}{\bibinfo{person}{Xiaonan Li}, \bibinfo{person}{Kai Lv}, \bibinfo{person}{Hang Yan}, \bibinfo{person}{Tianyang Lin}, \bibinfo{person}{Wei Zhu}, \bibinfo{person}{Yuan Ni}, \bibinfo{person}{Guotong Xie}, \bibinfo{person}{Xiaoling Wang}, {and} \bibinfo{person}{Xipeng Qiu}.} \bibinfo{year}{2023}\natexlab{b}.
\newblock \showarticletitle{Unified demonstration retriever for in-context learning}.
\newblock \bibinfo{journal}{\emph{arXiv preprint arXiv:2305.04320}} (\bibinfo{year}{2023}).
\newblock


\bibitem[Li and Qiu(2023)]%
        {li2023finding}
\bibfield{author}{\bibinfo{person}{Xiaonan Li} {and} \bibinfo{person}{Xipeng Qiu}.} \bibinfo{year}{2023}\natexlab{}.
\newblock \showarticletitle{Finding support examples for in-context learning}.
\newblock \bibinfo{journal}{\emph{arXiv preprint arXiv:2302.13539}} (\bibinfo{year}{2023}).
\newblock


\bibitem[Liang et~al\mbox{.}(2024)]%
        {liang2024survey}
\bibfield{author}{\bibinfo{person}{Zijing Liang}, \bibinfo{person}{Yanjie Xu}, \bibinfo{person}{Yifan Hong}, \bibinfo{person}{Penghui Shang}, \bibinfo{person}{Qi Wang}, \bibinfo{person}{Qiang Fu}, {and} \bibinfo{person}{Ke Liu}.} \bibinfo{year}{2024}\natexlab{}.
\newblock \showarticletitle{A Survey of Multimodel Large Language Models}. In \bibinfo{booktitle}{\emph{Proceedings of the 3rd International Conference on Computer, Artificial Intelligence and Control Engineering}}. \bibinfo{pages}{405--409}.
\newblock


\bibitem[Liu et~al\mbox{.}(2023)]%
        {liu2023visual}
\bibfield{author}{\bibinfo{person}{Haotian Liu}, \bibinfo{person}{Chunyuan Li}, \bibinfo{person}{Qingyang Wu}, {and} \bibinfo{person}{Yong~Jae Lee}.} \bibinfo{year}{2023}\natexlab{}.
\newblock \showarticletitle{Visual instruction tuning}.
\newblock \bibinfo{journal}{\emph{Advances in neural information processing systems}}  \bibinfo{volume}{36} (\bibinfo{year}{2023}), \bibinfo{pages}{34892--34916}.
\newblock


\bibitem[Liu et~al\mbox{.}(2024)]%
        {liu2024survey}
\bibfield{author}{\bibinfo{person}{Hanchao Liu}, \bibinfo{person}{Wenyuan Xue}, \bibinfo{person}{Yifei Chen}, \bibinfo{person}{Dapeng Chen}, \bibinfo{person}{Xiutian Zhao}, \bibinfo{person}{Ke Wang}, \bibinfo{person}{Liping Hou}, \bibinfo{person}{Rongjun Li}, {and} \bibinfo{person}{Wei Peng}.} \bibinfo{year}{2024}\natexlab{}.
\newblock \showarticletitle{A survey on hallucination in large vision-language models}.
\newblock \bibinfo{journal}{\emph{arXiv preprint arXiv:2402.00253}} (\bibinfo{year}{2024}).
\newblock


\bibitem[Liu et~al\mbox{.}(2022)]%
        {liu2022makes}
\bibfield{author}{\bibinfo{person}{Jiachang Liu}, \bibinfo{person}{Dinghan Shen}, \bibinfo{person}{Yizhe Zhang}, \bibinfo{person}{William~B Dolan}, \bibinfo{person}{Lawrence Carin}, {and} \bibinfo{person}{Weizhu Chen}.} \bibinfo{year}{2022}\natexlab{}.
\newblock \showarticletitle{What Makes Good In-Context Examples for GPT-3?}. In \bibinfo{booktitle}{\emph{Proceedings of Deep Learning Inside Out (DeeLIO 2022): The 3rd Workshop on Knowledge Extraction and Integration for Deep Learning Architectures}}. \bibinfo{pages}{100--114}.
\newblock


\bibitem[Mavromatis et~al\mbox{.}(2023)]%
        {mavromatis2023examples}
\bibfield{author}{\bibinfo{person}{Costas Mavromatis}, \bibinfo{person}{Balasubramaniam Srinivasan}, \bibinfo{person}{Zhengyuan Shen}, \bibinfo{person}{Jiani Zhang}, \bibinfo{person}{Huzefa Rangwala}, \bibinfo{person}{Christos Faloutsos}, {and} \bibinfo{person}{George Karypis}.} \bibinfo{year}{2023}\natexlab{}.
\newblock \showarticletitle{Which examples to annotate for in-context learning? towards effective and efficient selection}.
\newblock \bibinfo{journal}{\emph{arXiv preprint arXiv:2310.20046}} (\bibinfo{year}{2023}).
\newblock


\bibitem[Min et~al\mbox{.}(2022)]%
        {min2022rethinking}
\bibfield{author}{\bibinfo{person}{Sewon Min}, \bibinfo{person}{Xinxi Lyu}, \bibinfo{person}{Ari Holtzman}, \bibinfo{person}{Mikel Artetxe}, \bibinfo{person}{Mike Lewis}, \bibinfo{person}{Hannaneh Hajishirzi}, {and} \bibinfo{person}{Luke Zettlemoyer}.} \bibinfo{year}{2022}\natexlab{}.
\newblock \showarticletitle{Rethinking the role of demonstrations: What makes in-context learning work?}
\newblock \bibinfo{journal}{\emph{arXiv preprint arXiv:2202.12837}} (\bibinfo{year}{2022}).
\newblock


\bibitem[Pan(2023)]%
        {pan2023context}
\bibfield{author}{\bibinfo{person}{Jane Pan}.} \bibinfo{year}{2023}\natexlab{}.
\newblock \emph{\bibinfo{title}{What in-context learning “learns” in-context: Disentangling task recognition and task learning}}.
\newblock \bibinfo{thesistype}{Master's\ thesis}. \bibinfo{school}{Princeton University}.
\newblock


\bibitem[Peng et~al\mbox{.}(2025)]%
        {peng2025lmm}
\bibfield{author}{\bibinfo{person}{Yingzhe Peng}, \bibinfo{person}{Gongrui Zhang}, \bibinfo{person}{Miaosen Zhang}, \bibinfo{person}{Zhiyuan You}, \bibinfo{person}{Jie Liu}, \bibinfo{person}{Qipeng Zhu}, \bibinfo{person}{Kai Yang}, \bibinfo{person}{Xingzhong Xu}, \bibinfo{person}{Xin Geng}, {and} \bibinfo{person}{Xu Yang}.} \bibinfo{year}{2025}\natexlab{}.
\newblock \showarticletitle{Lmm-r1: Empowering 3b lmms with strong reasoning abilities through two-stage rule-based rl}.
\newblock \bibinfo{journal}{\emph{arXiv preprint arXiv:2503.07536}} (\bibinfo{year}{2025}).
\newblock


\bibitem[Qian et~al\mbox{.}(2024)]%
        {qian2024sub}
\bibfield{author}{\bibinfo{person}{Jian Qian}, \bibinfo{person}{Miao Sun}, \bibinfo{person}{Sifan Zhou}, \bibinfo{person}{Ziyu Zhao}, \bibinfo{person}{Ruizhi Hun}, {and} \bibinfo{person}{Patrick Chiang}.} \bibinfo{year}{2024}\natexlab{}.
\newblock \showarticletitle{Sub-SA: Strengthen In-context Learning via Submodular Selective Annotation}.
\newblock \bibinfo{journal}{\emph{arXiv preprint arXiv:2407.05693}} (\bibinfo{year}{2024}).
\newblock


\bibitem[Radford et~al\mbox{.}(2021)]%
        {radford2021learning}
\bibfield{author}{\bibinfo{person}{Alec Radford}, \bibinfo{person}{Jong~Wook Kim}, \bibinfo{person}{Chris Hallacy}, \bibinfo{person}{Aditya Ramesh}, \bibinfo{person}{Gabriel Goh}, \bibinfo{person}{Sandhini Agarwal}, \bibinfo{person}{Girish Sastry}, \bibinfo{person}{Amanda Askell}, \bibinfo{person}{Pamela Mishkin}, \bibinfo{person}{Jack Clark}, {et~al\mbox{.}}} \bibinfo{year}{2021}\natexlab{}.
\newblock \showarticletitle{Learning transferable visual models from natural language supervision}. In \bibinfo{booktitle}{\emph{International conference on machine learning}}. PmLR, \bibinfo{pages}{8748--8763}.
\newblock


\bibitem[Ren et~al\mbox{.}(2021)]%
        {ren2021survey}
\bibfield{author}{\bibinfo{person}{Pengzhen Ren}, \bibinfo{person}{Yun Xiao}, \bibinfo{person}{Xiaojun Chang}, \bibinfo{person}{Po-Yao Huang}, \bibinfo{person}{Zhihui Li}, \bibinfo{person}{Brij~B Gupta}, \bibinfo{person}{Xiaojiang Chen}, {and} \bibinfo{person}{Xin Wang}.} \bibinfo{year}{2021}\natexlab{}.
\newblock \showarticletitle{A survey of deep active learning}.
\newblock \bibinfo{journal}{\emph{ACM computing surveys (CSUR)}} \bibinfo{volume}{54}, \bibinfo{number}{9} (\bibinfo{year}{2021}), \bibinfo{pages}{1--40}.
\newblock


\bibitem[Russakovsky et~al\mbox{.}(2015)]%
        {russakovsky2015imagenet}
\bibfield{author}{\bibinfo{person}{Olga Russakovsky}, \bibinfo{person}{Jia Deng}, \bibinfo{person}{Hao Su}, \bibinfo{person}{Jonathan Krause}, \bibinfo{person}{Sanjeev Satheesh}, \bibinfo{person}{Sean Ma}, \bibinfo{person}{Zhiheng Huang}, \bibinfo{person}{Andrej Karpathy}, \bibinfo{person}{Aditya Khosla}, \bibinfo{person}{Michael Bernstein}, {et~al\mbox{.}}} \bibinfo{year}{2015}\natexlab{}.
\newblock \showarticletitle{Imagenet large scale visual recognition challenge}.
\newblock \bibinfo{journal}{\emph{International journal of computer vision}}  \bibinfo{volume}{115} (\bibinfo{year}{2015}), \bibinfo{pages}{211--252}.
\newblock


\bibitem[Sener and Savarese(2017)]%
        {sener2017active}
\bibfield{author}{\bibinfo{person}{Ozan Sener} {and} \bibinfo{person}{Silvio Savarese}.} \bibinfo{year}{2017}\natexlab{}.
\newblock \showarticletitle{Active learning for convolutional neural networks: A core-set approach}.
\newblock \bibinfo{journal}{\emph{arXiv preprint arXiv:1708.00489}} (\bibinfo{year}{2017}).
\newblock


\bibitem[Tsimpoukelli et~al\mbox{.}(2021)]%
        {tsimpoukelli2021multimodal}
\bibfield{author}{\bibinfo{person}{Maria Tsimpoukelli}, \bibinfo{person}{Jacob~L Menick}, \bibinfo{person}{Serkan Cabi}, \bibinfo{person}{SM Eslami}, \bibinfo{person}{Oriol Vinyals}, {and} \bibinfo{person}{Felix Hill}.} \bibinfo{year}{2021}\natexlab{}.
\newblock \showarticletitle{Multimodal few-shot learning with frozen language models}.
\newblock \bibinfo{journal}{\emph{Advances in Neural Information Processing Systems}}  \bibinfo{volume}{34} (\bibinfo{year}{2021}), \bibinfo{pages}{200--212}.
\newblock


\bibitem[Wang et~al\mbox{.}(2024)]%
        {wang2024qwen2}
\bibfield{author}{\bibinfo{person}{Peng Wang}, \bibinfo{person}{Shuai Bai}, \bibinfo{person}{Sinan Tan}, \bibinfo{person}{Shijie Wang}, \bibinfo{person}{Zhihao Fan}, \bibinfo{person}{Jinze Bai}, \bibinfo{person}{Keqin Chen}, \bibinfo{person}{Xuejing Liu}, \bibinfo{person}{Jialin Wang}, \bibinfo{person}{Wenbin Ge}, {et~al\mbox{.}}} \bibinfo{year}{2024}\natexlab{}.
\newblock \showarticletitle{Qwen2-vl: Enhancing vision-language model's perception of the world at any resolution}.
\newblock \bibinfo{journal}{\emph{arXiv preprint arXiv:2409.12191}} (\bibinfo{year}{2024}).
\newblock


\bibitem[Welinder et~al\mbox{.}(2010)]%
        {welinder2010caltech}
\bibfield{author}{\bibinfo{person}{Peter Welinder}, \bibinfo{person}{Steve Branson}, \bibinfo{person}{Takeshi Mita}, \bibinfo{person}{Catherine Wah}, \bibinfo{person}{Florian Schroff}, \bibinfo{person}{Serge Belongie}, {and} \bibinfo{person}{Pietro Perona}.} \bibinfo{year}{2010}\natexlab{}.
\newblock \showarticletitle{Caltech-UCSD birds 200}.
\newblock  (\bibinfo{year}{2010}).
\newblock


\bibitem[Yang et~al\mbox{.}(2023a)]%
        {yang2023lever}
\bibfield{author}{\bibinfo{person}{Xu Yang}, \bibinfo{person}{Yingzhe Peng}, \bibinfo{person}{Haoxuan Ma}, \bibinfo{person}{Shuo Xu}, \bibinfo{person}{Chi Zhang}, \bibinfo{person}{Yucheng Han}, {and} \bibinfo{person}{Hanwang Zhang}.} \bibinfo{year}{2023}\natexlab{a}.
\newblock \showarticletitle{Lever LM: Configuring In-Context Sequence to Lever Large Vision Language Models}.
\newblock \bibinfo{journal}{\emph{arXiv e-prints}} (\bibinfo{year}{2023}), \bibinfo{pages}{arXiv--2312}.
\newblock


\bibitem[Yang et~al\mbox{.}(2023b)]%
        {yang2023exploring}
\bibfield{author}{\bibinfo{person}{Xu Yang}, \bibinfo{person}{Yongliang Wu}, \bibinfo{person}{Mingzhuo Yang}, \bibinfo{person}{Haokun Chen}, {and} \bibinfo{person}{Xin Geng}.} \bibinfo{year}{2023}\natexlab{b}.
\newblock \showarticletitle{Exploring diverse in-context configurations for image captioning}.
\newblock \bibinfo{journal}{\emph{Advances in Neural Information Processing Systems}}  \bibinfo{volume}{36} (\bibinfo{year}{2023}), \bibinfo{pages}{40924--40943}.
\newblock


\bibitem[Ye et~al\mbox{.}(2023)]%
        {ye2023mplug}
\bibfield{author}{\bibinfo{person}{Qinghao Ye}, \bibinfo{person}{Haiyang Xu}, \bibinfo{person}{Guohai Xu}, \bibinfo{person}{Jiabo Ye}, \bibinfo{person}{Ming Yan}, \bibinfo{person}{Yiyang Zhou}, \bibinfo{person}{Junyang Wang}, \bibinfo{person}{Anwen Hu}, \bibinfo{person}{Pengcheng Shi}, \bibinfo{person}{Yaya Shi}, {et~al\mbox{.}}} \bibinfo{year}{2023}\natexlab{}.
\newblock \showarticletitle{mplug-owl: Modularization empowers large language models with multimodality}.
\newblock \bibinfo{journal}{\emph{arXiv preprint arXiv:2304.14178}} (\bibinfo{year}{2023}).
\newblock


\bibitem[You et~al\mbox{.}(2022)]%
        {you2022end}
\bibfield{author}{\bibinfo{person}{Chenyu You}, \bibinfo{person}{Nuo Chen}, \bibinfo{person}{Fenglin Liu}, \bibinfo{person}{Shen Ge}, \bibinfo{person}{Xian Wu}, {and} \bibinfo{person}{Yuexian Zou}.} \bibinfo{year}{2022}\natexlab{}.
\newblock \showarticletitle{End-to-end spoken conversational question answering: Task, dataset and model}.
\newblock \bibinfo{journal}{\emph{arXiv preprint arXiv:2204.14272}} (\bibinfo{year}{2022}).
\newblock


\bibitem[You et~al\mbox{.}(2021a)]%
        {you2021mrd}
\bibfield{author}{\bibinfo{person}{Chenyu You}, \bibinfo{person}{Nuo Chen}, {and} \bibinfo{person}{Yuexian Zou}.} \bibinfo{year}{2021}\natexlab{a}.
\newblock \showarticletitle{MRD-Net: Multi-Modal Residual Knowledge Distillation for Spoken Question Answering.}. In \bibinfo{booktitle}{\emph{IJCAI}}. \bibinfo{pages}{3985--3991}.
\newblock


\bibitem[You et~al\mbox{.}(2021b)]%
        {you2021self}
\bibfield{author}{\bibinfo{person}{Chenyu You}, \bibinfo{person}{Nuo Chen}, {and} \bibinfo{person}{Yuexian Zou}.} \bibinfo{year}{2021}\natexlab{b}.
\newblock \showarticletitle{Self-supervised Contrastive Cross-Modality Representation Learning for Spoken Question Answering}. In \bibinfo{booktitle}{\emph{Findings of the Association for Computational Linguistics: EMNLP}}.
\newblock


\bibitem[Zhai et~al\mbox{.}(2023)]%
        {zhai2023sigmoid}
\bibfield{author}{\bibinfo{person}{Xiaohua Zhai}, \bibinfo{person}{Basil Mustafa}, \bibinfo{person}{Alexander Kolesnikov}, {and} \bibinfo{person}{Lucas Beyer}.} \bibinfo{year}{2023}\natexlab{}.
\newblock \showarticletitle{Sigmoid loss for language image pre-training}. In \bibinfo{booktitle}{\emph{Proceedings of the IEEE/CVF international conference on computer vision}}. \bibinfo{pages}{11975--11986}.
\newblock


\bibitem[Zhang et~al\mbox{.}(2024)]%
        {zhang2024visually}
\bibfield{author}{\bibinfo{person}{Yuhui Zhang}, \bibinfo{person}{Alyssa Unell}, \bibinfo{person}{Xiaohan Wang}, \bibinfo{person}{Dhruba Ghosh}, \bibinfo{person}{Yuchang Su}, \bibinfo{person}{Ludwig Schmidt}, {and} \bibinfo{person}{Serena Yeung-Levy}.} \bibinfo{year}{2024}\natexlab{}.
\newblock \showarticletitle{Why are visually-grounded language models bad at image classification?}
\newblock \bibinfo{journal}{\emph{arXiv preprint arXiv:2405.18415}} (\bibinfo{year}{2024}).
\newblock


\bibitem[Zhang et~al\mbox{.}(2023)]%
        {zhang2023makes}
\bibfield{author}{\bibinfo{person}{Yuanhan Zhang}, \bibinfo{person}{Kaiyang Zhou}, {and} \bibinfo{person}{Ziwei Liu}.} \bibinfo{year}{2023}\natexlab{}.
\newblock \showarticletitle{What makes good examples for visual in-context learning?}
\newblock \bibinfo{journal}{\emph{Advances in Neural Information Processing Systems}}  \bibinfo{volume}{36} (\bibinfo{year}{2023}), \bibinfo{pages}{17773--17794}.
\newblock


\bibitem[Zhu et~al\mbox{.}(2023)]%
        {zhu2023minigpt}
\bibfield{author}{\bibinfo{person}{Deyao Zhu}, \bibinfo{person}{Jun Chen}, \bibinfo{person}{Xiaoqian Shen}, \bibinfo{person}{Xiang Li}, {and} \bibinfo{person}{Mohamed Elhoseiny}.} \bibinfo{year}{2023}\natexlab{}.
\newblock \showarticletitle{Minigpt-4: Enhancing vision-language understanding with advanced large language models}.
\newblock \bibinfo{journal}{\emph{arXiv preprint arXiv:2304.10592}} (\bibinfo{year}{2023}).
\newblock


\end{thebibliography}

\appendix

\section{Implementation Details}
\label{sec:impl_details}

\subsection{Coreset Initialization}
\label{subsec:coreset_init}

\noindent \textbf{K-center initialization. }
K-center-greedy algorithm is widely used in active learning with the aim of selecting the most representative samples for labeling. As described in Algorithm~\ref{alg:greedy}, its core concept involves iteratively choosing the point that is furthest from the current set of centers as a new center, thereby ensuring that the selected set of center points can cover all other points. Given our coreset configuration, which maintains an equal number of samples for each class. To achieve this, we slightly modify the original algorithm to obtain a sample set for each category, with the objective of maximizing the diversity within each class.
 
\begin{algorithm}
\caption{Modified K-Center Greedy Algorithm for Multi-Class Support set}
\label{alg:greedy}
\begin{algorithmic}
\State \textbf{Input:} A support set \( S \), the number of classes \( j \), the total number of centers \( m \)
\State \textbf{Output:} A set of \( m \) centers \( C \) with \( \frac{m}{j} \) centers from each class
\State Initialize \( C = \emptyset \)
\State Let \( S_1, S_2, \dots, S_j \) be the subsets of \( S \) corresponding to each class
\For{\( i = 1 \) \textbf{to} \( j \)}
    \State Let \( m_i = \frac{m}{j} \) be the number of centers to select from class \( i \)
    \State Initialize \( C_i = \emptyset \)
    \State Randomly select a point \( p \in S_i \) and set \( C_i = \{ p \} \)
    \For{\( k = 2 \) \textbf{to} \( m_i \)}
        \State Initialize \( \text{max\_distance} = 0 \)
        \State Initialize \( \text{new\_center} = \text{None} \)
        \For{\( p \in S_i \)}
            \State Compute \( \text{min\_distance} = \min_{c \in C_i} \text{distance}(p, c) \)
            \If{\( \text{min\_distance} > \text{max\_distance} \)}
                \State Set \( \text{max\_distance} = \text{min\_distance} \)
                \State Set \( \text{new\_center} = p \)
            \EndIf
        \EndFor
        \State Add \( \text{new\_center} \) to \( C_i \)
    \EndFor
    \State Add all points in \( C_i \) to \( C \)
\EndFor
\State \Return \( C \)
\label{al}
\end{algorithmic}
\end{algorithm}

\noindent \textbf{Infoscore initialization. }
In text classifcation tasks, the Infoscore of a sample $e=\{x,y\} $ is calculated as follows:
\begin{align}
I(e, \mathcal{D}) &= \sum_{e' \in \mathcal{D}} c(e, e') \label{eq:IeD} \\
c(e, e') &= p_G(y' \mid x, y, x') - p_G(y' \mid x'), \label{eq:ce}
\end{align}
where $e' = \{x', y'\}$, $G$ is LM, and $D$ is the training dataset. Equation~(\ref{eq:ce}) is the gap between the probabilities of the ground truth $y'$ conditioned on $(e, x')$ and $(x')$, respectively. So it evaluates how informative $e$ is for the LM to correctly classify $x'$ and thus measures $e$'s contribution for $e'$ in ICL. Hence, $I(e, D)$, the sum of Equation~(\ref{eq:ce}) over D, can evaluate the example’s in-context informativeness. 

We adapt this method for the image classification task. Similar to the formula mentioned above, the difference is that we replace LM with LVLM, which is used in the experiments section. For our experiment, the Infoscore of a sample 
$e=\{x,y\} $ is calculated as follows:
\begin{align}
I(e, \mathcal{S}) &= \sum_{e' \in \mathcal{S}} c(e, e') \label{eq:IeD_i} \\
c(e, e') &= p_G(y' \mid x, y, x') - p_G(y' \mid x'), \label{eq:ce_i}
\end{align}
where $e' = \{x', y'\}$, $x'$ is image and $y'$ is label, $G$ is LVLM, and $S$ is the support set. Equation~(\ref{eq:ce_i}) is the gap between the probabilities of the ground truth $y'$ conditioned on $(e, x')$ and $(x')$, respectively. So it evaluates how informative $e$ is for the LVLM to correctly classify $x'$ and thus measures $e$'s contribution for $e'$ in ICL. Hence, $I(e, S)$, the sum of Equation~(\ref{eq:ce_i}) over $S$, can evaluate the example’s in-context informativeness.

\subsection{Datasets}
\label{subsec:datasets}
\noindent \textbf{CUB-200 }
 contains a total of 11,788 bird images, encompassing 200 bird subclasses. The training set consists of 5,994 images, and the validation set comprises 5,794 images. Each image is provided with image class label information.

\noindent \textbf{Stanford Dogs }
contains 20,580 images of 120 breeds of dogs from around the world. This dataset has been built using images and annotation from ImageNet for the task of fine-grained image categorization.

\noindent \textbf{ImageNet-100 }
is the subset of ImageNet dataset which contains images of 100 classes from the original dataset. It consists of total 130,000 images for training set (1,300 images per class) and total 5,000 images for validation set (50 images per class).

\subsection{Ablation Study on Hyperparameters}
\label{subsec:parameters}

\begin{table*}
\centering
\begin{tabular}{c|cccccccccc}
\toprule%
Model   & 0.1    & 0.2  & 0.3   & 0.4  & 0.5 & 0.6  & 0.7 & 0.8 & 0.9 & 1.0\\ 
\midrule%
OF-3B    & 72.30 & \textbf{76.96} & 76.76& 76.70 & 76.06 & 74.72 & 73.88 & 72.50 & 71.72 & 70.92\\
IDE-8B   & 90.50 & \textbf{91.58} & 90.80& 91.14& 91.04 & 91.16 & 91.38 & 91.48 & 90.78  &  90.60     \\
\bottomrule%
\end{tabular}%
\caption{2-shot ICL perforamce ccuracy (\%) of OF-3B and
IDE-8B on CUB-200 under different update rate $\alpha$. Results are evaluated under KeCO-DS in online scenario.}
\label{tab: different_alpha}
\end{table*}

\begin{table}
\centering
\begin{tabular}{c|ccccccc}
\toprule%
Epoch   & 200    & 600  & 800   & 1000  & 2000 & 3000 & 4000  \\ 
\midrule%
1    & 68.28    & 63.27  & 62.68   & 60.68  & 54.69  & 51.36& 51.81 \\
5   & 73.90    & 74.65  & 74.44   & 74.47  & 74.45 & 68.48 & 69.81     \\
10   & 74.02    & 74.54  & 74.37   & 74.20  & 74.49 & 73.36 & 74.78     \\
15   & 74.73    & 74.09  & 73.97   & 74.73  & 73.67  & 73.25 & 74.15     \\
\bottomrule%
\end{tabular}%
\caption{2-shot ICL perforamce ccuracy (\%) of OF-3B  on CUB-200 under epoch $e$ and batch size $b$. Results are evaluated under KeCO-DS in general scenario.}
\label{tab: epoch}
\end{table}

We conducted an investigation into the impact of three hyperparameters, namely the update rate $r$, epoch $e$, and batch size $b$ on the ICL performance. First, we explored how the 2-shot performance of OF-3B and IDE-8B under the KeCO-DS setting on the CUB-200 dataset changes with varying $r$. Table~\ref{tab: different_alpha} demonstrates the trends in accuracy as $r$ changes from 0.1 to 1.0. As depicted, KeCO-DS shows an increasing trend from 0.0 to 0.2, achieving optimal performance at $r=0.2$, and then gradually declining from 0.2 to 1.0. This suggests that an update rate of 0.2 is generally well-suited for balancing stability and adaptability. It further implies that higher update rates may lead to the destabilization of the coreset updating process, making it less effective at retaining relevant knowledge while adapting to new information.



Next, we explore the impact of the number of epoch $e$ and the batch size $b$ on ICL performance. As can be seen from Table~\ref{tab: epoch}, when $e=1$, varying the batch size from 200 to 4,000 results in lower performance compared to when the number of epochs is set to 5, 10, or 15. This highlights the effectiveness of repeatedly using untapped data to update the coreset. However, when $e$ increases (e.g. to 10, 15), the variation in $b$ does not significantly impact the results, suggesting that the role of $b$ becomes less important when untapped data can be reused. Interestingly, when $e=1$, we find that a larger $b$ actually leads to worse performance. The smaller $b$ allow for more frequent updates, resulting in a greater number of update iterations. Since each update serves to retain category-relevant information and blur category-irrelevant information, a higher number of updates means more opportunities to enhance the information in the coreset, especially in online scenario where untapped data can only be utilized once. Furthermore, when $e=1$ and $b=1$, this is equivalent to generalizing KeCO to a specific online scenario.

\subsection{Analyses of the Size of Additional Data and
Coreset}

\begin{table}
	\setlength{\tabcolsep}{4pt}
	\centering
	\begin{tabular}{c c ccc c ccc} 
		\toprule
		\multirow{2}{*}{\textbf{Method}} &\multirow{2}{*}{}& \multicolumn{3}{c}{OF-3B}   &  & \multicolumn{3}{c}{IDE-8B}  \\
		&                   & 1:2 & 1:4 & 1:6 & &  1:2  & 1:4 & 1:6 \\ 
		\cline{1-1}\cline{3-5}\cline{7-9}
		FS-IC    &  & 49.80   & 49.80 & 49.80   &  & 74.42 & 74.42  &74.42 \\
		FS-IS    &  & 58.67   & 61.42   & 63.42  &  & 75.68 & 76.89  & 77.45 \\
        KeCO-RD     &  & 69.49  & 72.20 & 73.58  &    & 78.32 & 79.66   & 79.13\\
        KeCO-DS     &  & 69.99  & 73.75 & 74.58  &    & 79.02 & 79.72 & 79.92   \\
        \hline

\textit{Online  Scenario}                 &   &  &  &   &  \\ \hline       
		KeCO-RS  &   & 63.02  & 67.06 & 69.93 &   & 77.30 & 78.38 & 78.66  \\
		KeCO-DS  &   & 65.56  & 69.92 &  72.47  &  & 77.33 & 78.87 & 79.52\\ \hline

	\end{tabular}
	\caption{2-shot ICL Performance accuracy (\%) of OF-3B and IDE-8B on Stanford Dogs under untapped set of different size. The coreset size is consistently set to 1200, and the ratio represents the comparison of the coreset size to the size of the untapped set. For instance, a ratio of 1:2 implies the size of the untapped set is 2400. Results are evaluated across two KeCO methods (KeCO-RS and KeCO-DS) and two baselines (FS-IC and FS-IS).}
	\label{tab: additional_data}
\end{table}

\begin{table}
	\setlength{\tabcolsep}{4pt}
	\centering
	\begin{tabular}{c c ccc c ccc} 
		\toprule
		\multirow{2}{*}{\textbf{Method}} &\multirow{2}{*}{}& \multicolumn{3}{c}{OF-3B}   &  & \multicolumn{3}{c}{IDE-8B}  \\
		&                   & 400 & 800 & 1000 & &  400 & 800  & 1000 \\ 
		\cline{1-1}\cline{3-5}\cline{7-9}
		FS-IC    &  & 33.53   & 44.65 & 48.11 &   & 82.84 & 83.59 & 84.65 \\
		FS-IS    &  & 61.04   & 61.04   & 61.04 &   & 85.62 & 85.62 & 85.62   \\
        KeCO-RS     &  & 69.59  & 71.85 & 72.90  &    & 85.05 & 85.33  & 85.46  \\
        KeCO-DS     &  & 71.83  & 73.21 & 74.20  &    & 85.67 & 85.90 & 87.38  \\
        \hline

\textit{Online  Scenario}                 &   &  &  &   &  \\ \hline       

		KeCO-RS  &   & 65.75  & 67.86 & 65.90 &   & 84.04 & 84.51 & 85.69  \\
		KeCO-DS  &   & 69.47  & 71.92 &  70.61  & & 85.38 & 86.69 & 85.91  \\ \hline
	\end{tabular}
	\caption{2-shot ICL Performance accuracy (\%) of OF-3B and IDE-8B on CUB-200 under coresets of different size . The support set size  is consistently set to 5000. For instance, if the coreset size is 800, then the size of the untapped set would be 5000 - 800 = 4200. Results are evaluated across two KeCO methods (KeCO-RS and KeCO-DS) and two baselines (FS-IC and FS-IS).}
	\label{tab: coreset}
\end{table}

From Table~\ref{tab: additional_data}, it can be observed that as the amount of untapped data increases, there is an improvement in the FS-IS, KeCO-RS, and KeCO-DS settings. For instance, when the ratio of coreset size to untapped data size increases from 1:2 to 1:6, the performance of OF-3B under the FS-IS setting improves from 58.67\% to 63.42\%, and IDE-8B under the same setting improves from 75.68\% to 77.45\%. Similarly, under the KeCO framework, when the ratio changes from 1:2 to 1:6, OF-3B's ICL performance under the KeCO-DS setting improves from 69.99\% to 74.58\%, and IDE-8B under the same setting improves from 79.02\% to 79.92\%.

As Table~\ref{tab: coreset} shows, increasing the coreset size from 400 to 1000 enhances the ICL performance of both OF-3B and IDE-8B under the FS-IC setting, as expected due to the coreset's increased information. Notably, in the online scenario, the ICL performance of the two LVLMs is higher with a coreset size of 800 and an untapped data size of 4200, compared to a coreset size of 1000 with an untapped data size of 4000. This could be because coreset sizes of 800 and 1000 provide 4 and 5 samples per class respectively, sufficient for 2-shot ICL. Therefore, when untapped data can only be utilized once, it might be more beneficial to update the coreset with more data than to simply have a larger initial coreset.


\end{document}